\documentclass[final,12pt,authoryear]{elsarticle}

\usepackage{amsmath, amsfonts,amssymb}
\usepackage{nicefrac}
\usepackage{pifont}
\usepackage{subcaption}
\usepackage{tikz}
\usepackage{pgfplots}
\usepackage{graphicx}
\usepackage{algorithm}
\usepackage{algorithmic}
\usepackage{multirow}

\usetikzlibrary{external}
\newcommand{\cmark}{\ding{51}}%
\newcommand{\xmark}{\ding{55}}%

\newtheorem{theorem}{Theorem}
\newtheorem{proposition}[theorem]{Proposition}
\newproof{proof}{Proof}
\newtheorem{lemma}[theorem]{Lemma}

\journal{Neural Networks}

\begin{document}

\begin{frontmatter}



\title{Low-Variance Forward Gradients using Direct Feedback Alignment and Momentum}


\author[1]{Florian Bacho \corref{cor1}}
\ead{f.bacho@kent.ac.uk}
\cortext[cor1]{Corresponding author}

\author[1]{Dominique Chu}
\ead{d.f.chu@kent.ac.uk}

\address[1]{CEMS, School of Computing, University of Kent, Canterbury, United Kingdom}

\begin{abstract}
	Supervised learning in deep neural networks is commonly performed using error backpropagation. However, the sequential propagation of errors during the backward pass limits its scalability and applicability to low-powered neuromorphic hardware. Therefore, there is growing interest in finding local alternatives to backpropagation. Recently proposed methods based on forward-mode automatic differentiation suffer from high variance in large deep neural networks, which affects convergence. In this paper, we propose the Forward Direct Feedback Alignment algorithm that combines Activity-Perturbed Forward Gradients with Direct Feedback Alignment and momentum. We provide both theoretical proofs and empirical evidence that our proposed method achieves lower variance than forward gradient techniques. In this way, our approach enables faster convergence and better performance when compared to other local alternatives to backpropagation and opens a new perspective for the development of online learning algorithms compatible with neuromorphic systems.
\end{abstract}


\end{frontmatter}

\section{Introduction}

Over the past decades, the Backpropagation (BP) algorithm \citep{error_backpropagation} has emerged as a crucial technique for training Deep Neural Networks (DNNs). However, despite its success, BP has limitations that restrict its efficiency and scalability.
\par
By sequentially propagating errors through multiple layers, BP limits the ability to parallelize the backward pass; this is often referred to as \textit{Backward Locking} \citep{random_dfa, dfa_scales_to_modern_dl, decoupled_parallel_bp} and leads to time-consuming gradient computations. Secondly, BP relies on the transport of symmetric weights during the backward pass. Known as the \textit{weight transport problem} \citep{deep_learning_without_weight_transport, random_fa, random_dfa}, it is a source of significant power consumption on dedicated neural processors \citep{sparse_dfa, dfa_scales_to_modern_dl, dfa_cnn_online_learning_processor, energy_efficient_dnn_learning_processor} and represents a major obstacle to the implementation of BP on low-powered continuous-time neuromorphic hardware \citep{event_driven_random_bp_neuromorphic}. Therefore, there is a growing need for alternatives to BP that can parallelize gradient computations without requiring global knowledge of the entire network.
\par
In this context, several alternatives to BP have been proposed \citep{deep_infomax, greedy_infomax, greedy_layerwise_learning, local_error_signals, deep_supervised_learning_using_local_errors, synthetic_gradients, hebbian_learning_meets_cnn, forward_forward, weight_perturbation, gemini, flipout, gradients_without_bp, learning_by_directional_gd, scaling_forward_gradient_with_local_losses, random_fa, random_dfa, learning_connections_in_dfa, binary_dfa, sparse_dfa}. For example, greedy learning can be used to sequentially learn complex representations of inputs \citep{greedy_infomax, forward_forward, greedy_layerwise_learning} or decoupled neural interfaces can be constructed to predict future gradients in an asynchronous manner \citep{synthetic_gradients}. Some approaches also introduce biologically inspired phenomenas such as Self-Backpropagation (SBP) that alternates between local unsupervised \textit{sleep} phases and non-local supervised \textit{wake} phases which reduces the computational cost of training \citep{biologically_plausible_reward_propagation}. Another example of BP alternative is the Random Feedback Alignment (FA) algorithm \citep{random_fa}. FA overcomes the weight transport problem by using random feedback weights for error backpropagation. Alternatively, the Random Direct Feedback Alignment (DFA) algorithm \citep{random_dfa} directly projects output errors onto hidden neurons using fixed linear random feedback connections, removing the need for sequential propagation of errors and allowing parallel gradient computation. Other approaches such as Direct Random Target Propagation (DRTP) directly projects targets instead of output errors onto hidden layers to  decrease the computational cost of DFA \citep{direct_target_propagation, biologically_plausible_reward_propagation}.
However, while FA, DFA and DRTP keep the feedback connections fixed, feedback learning mechanisms can be introduced to improve the performance, as in the recently proposed Direct Kolen-Pollack (DKP) algorithm \citep{learning_connections_in_dfa}.
\par
Methods based on Forward-Mode Automatic Differentiation have recently gained attention \citep{review_ad}. Referred to as \textit{Forward Gradients} \citep{gradients_without_bp, learning_by_directional_gd, scaling_forward_gradient_with_local_losses}, these techniques evaluate directional derivatives in random directions during the forward pass to compute unbiased gradient estimates without backpropagation. However, in large DNNs, forward gradients suffer from high variance which has a detrimental effect on convergence. \citep{scaling_forward_gradient_with_local_losses, learning_by_directional_gd}. While stochastic gradient descent (SGD) is guaranteed to converge for unbiased gradients estimators with sufficiently small learning rates \citep{stochastic_approximation_method}, theoretical analyses have shown that the convergence rate of SGD depends on the variance of the estimates \citep{convex_optimization, optimization_methods_large_scale_ml, statistical_study_of_online_learning, analysis_of_stochastic_approximation_algorithms, sdg_weighted_sampling, convergence_sdg, sdg_general_analysis, gradient_variance_deep_learning}.
Estimators with low variance have less variability and are more consistent, leading to better convergence than those with high variance.
\par
One approach for reducing the variance of Forward Gradients is to perturb neuron activations instead of weights \citep{scaling_forward_gradient_with_local_losses}. Because DNNs typically have fewer neurons than weights, perturbing neurons results in estimating fewer derivatives through forward gradients, leading to lower variance. Additionally, local greedy loss functions can be used to further reduce the variance, where each loss function trains only a small portion of the network. As demonstrated by \citep{scaling_forward_gradient_with_local_losses}, this approach improves the convergence of Local MLP Mixers over the original FG algorithm. However, the method circumvents large gradient variance by ensuring that each local loss function only trains a small number of neurons. In the case where layers contain large numbers of neurons, the method would still suffer from high variance. Therefore, alternative solutions should be found to better reduce the variance of forward gradients in DNNs.
\par
In this work, we propose the Forward Direct Feedback Alignment (FDFA) algorithm, a method that combines Activity-Perturbed Forward Gradients \citep{scaling_forward_gradient_with_local_losses} with Direct Feedback Alignment \citep{random_dfa} and momentum to computing low-variance gradient estimates. Our method addresses the limitations of BP by avoiding sequential error propagation and the transport of weights. We present theoretical and empirical results that demonstrate the effectiveness of FDFA in reducing the variance of gradient estimates, enabling fast convergence with DNNs. Compared to other Forward Gradient and Direct Feedback Alignment methods, FDFA achieves better performance with both fully-connected and convolutional neural networks, making it a promising alternative for scalable and energy-efficient training of DNNs.

\section{Background}

We start by reviewing the Forward Gradient algorithm \citep{gradients_without_bp} and its DNN applications \citep{learning_by_directional_gd, scaling_forward_gradient_with_local_losses} as well as Direct Feedback Alignment \citep{random_dfa}, which form the technical foundation of the FDFA algorithm.

\subsection{Forward Gradient (FG)}

\begin{figure}[!tbp]
	\centering
	\begin{subfigure}[t]{0.35\linewidth}
		\includegraphics[width=\linewidth]{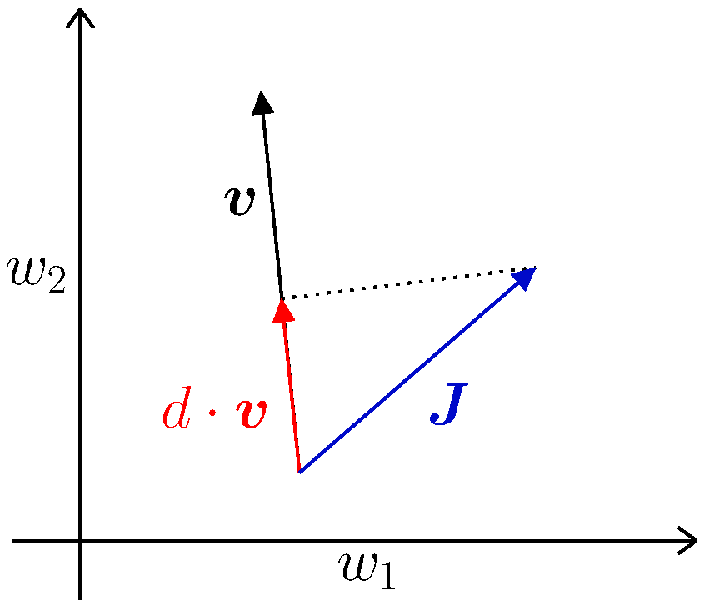}
		\caption{}
		\label{fig:dir_der_projection}
	\end{subfigure}%
	\begin{subfigure}[t]{0.35\linewidth}
		\includegraphics[width=\linewidth]{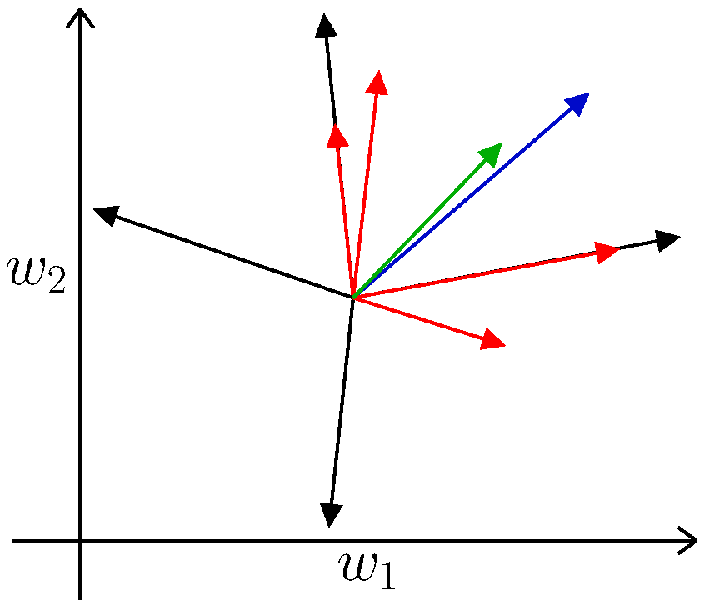}
		\caption{}
		\label{fig:forward_gradient}
	\end{subfigure}
	\caption{Figure \ref{fig:dir_der_projection}: Projection of the Jacobian $J$ at $\boldsymbol{w}$ onto a given direction $\boldsymbol{v}$. The vector $d \cdot \boldsymbol{v}$ is obtained by scaling the direction $\boldsymbol{v}$ by the directional derivative $d$ evaluated at $\boldsymbol{w}$ in the direction of $\boldsymbol{v}$. Figure \ref{fig:forward_gradient}: The expected directional derivative (green arrow), computed by averaging directional gradients (red arrows) over many random directions (black arrows), is an unbiased estimate of the true gradient (blue arrow).}
	\label{fig:directional_derivative_gradient}
\end{figure}

The Forward Gradient (FG) algorithm \citep{gradients_without_bp, learning_by_directional_gd} is a recent weight perturbation technique that uses Forward-Mode Automatic Differentiation (AD) \citep{review_ad} to estimate gradients without backpropagation. Consider a differentiable function $\boldsymbol{f} : \mathbb{R}^m \mapsto \mathbb{R}^n$ and a vector $\boldsymbol{v} \in \mathbb{R}^{m}$, Forward-Mode AD evaluates the directional gradient $\boldsymbol{d}=\boldsymbol{J} \cdot \boldsymbol{v}$ of $\boldsymbol{f}$ in the direction $\boldsymbol{v}$. Here, $\boldsymbol{J} \in \mathbb{R}^{n \times m}$ is the Jacobian matrix of $f$, and $\boldsymbol{d}$ is obtained by computing the matrix-vector product between $\boldsymbol{J}$ and $\boldsymbol{v}$ during the function evaluation.
\par
By sampling each element of the direction vector $v_{i} \sim \mathcal{N}\left(0, 1\right)$ from a standard normal distribution and multiplying back each computed directional derivatives by $\boldsymbol{v}$, an unbiased estimate of the Jacobian is computed:
\begin{equation}
	\mathbb{E}\left[ \boldsymbol{d} \otimes \boldsymbol{v}\right] = \mathbb{E}\left[ \left(\boldsymbol{J} \cdot \boldsymbol{v}\right) \otimes \boldsymbol{v}\right] = \boldsymbol{J}
\end{equation}
Here, $\otimes$ is the outer product. See \citep{gradients_without_bp} or Theorem \ref{the:forward_derivatives_unbiasedness} in Appendix for the proof of unbiasedness and Figure \ref{fig:directional_derivative_gradient} for a visual representation of forward gradients.

\subsection{Weight-Perturbed Forward Gradient (FG-W)}

When the FG algorithm is applied to DNNs, a random perturbation matrix $\boldsymbol{V}^{(l)} \in \mathbb{R}^{n_l \times n_{l-1}}$ is drawn from a standard normal distribution for each weight matrix $\boldsymbol{W} \in \mathbb{R}^{n_l \times n_{l-1}}$ of each layer $l \leq L$. Forward-Mode AD thus computes the directional derivative $d^\mathrm{FG-W}$ of the loss $\mathcal{L}\left(\boldsymbol{x}\right)$ along the drawn perturbation, such as:
\begin{equation}
	d^\mathrm{FG-W} = \sum_{l=1}^{L} \sum_{i=1}^{n_l} \sum_{j=1}^{n_{l-1}} \frac{\partial \mathcal{L}\left(\boldsymbol{x}\right)}{\partial w^{(l)}_{i,j}} v^{(l)}_{i,j}
\end{equation}
where $v^{(l)}_{i,j}$ is the element of the perturbation matrix $V^{(l)}$ in row $i$ and column $j$, associated with the weight $w^{(l)}_{i,j}$.
\par
Thus refered to as Weight-Perturbed Forward Gradient (FG-W) \citep{scaling_forward_gradient_with_local_losses}, the gradient estimate $g^\mathrm{FG-W}(\boldsymbol{W}^{(l)})$ for the weights of the layer $l$ is obtained by scaling its perturbation $\boldsymbol{V}^{(l)}$ matrix with the directional derivative $d^\mathrm{FG-W}$:
\begin{equation}
	g^\mathrm{FG-W}\left(\boldsymbol{W}^{(l)}\right) := d^\mathcal{\mathrm{FG-W}} \boldsymbol{V}^{(l)}
\end{equation}
See Algorithm \ref{alg:forward_gradient} in Appendix for the full algorithm applied to a fully-connected DNN.
\par
It is essential to understand that the directional derivative $d^\mathrm{FG-W}$, takes the form of a \textit{scalar}. This scalar is defined as the result of the vector product between the true gradient and the given direction that is implicitely evaluated with Forward-Mode AD. Unlike Reverse-Mode AD, where the individual explicit derivatives constituting the gradient are accessible, Forward-Mode AD does not provide them directly. Therefore, multiplying the directional derivative by the given direction is an essential step for computing an unbiased approximation of the gradient. For more details, refer to \citep{gradients_without_bp}.
\par
However, it has been previously shown that the variance of FG-W scales poorly with the number of parameters in DNNs which impacts the convergence of SGD \citep{scaling_forward_gradient_with_local_losses}.

\subsection{Activity-Perturbed Forward Gradient (FG-A)}

To address the variance issues of FG-W, \citeauthor{scaling_forward_gradient_with_local_losses} proposed the \textit{Activity-Perturbed FG} (FG-A) algorithm that perturbs neurons activations instead of weights. A perturbation vector $\boldsymbol{u}^{(l)} \in \mathbb{R}^{n_l}$ is drawn from a multivariate standard normal distribution for each layer $l \leq L$. Forward-Mode AD thus computes the following directional derivative:
\begin{equation}
	d^\mathrm{FG-A} = \sum_{l=1}^{L} \sum_{i=1}^{n_l} \frac{\partial \mathcal{L}\left(\boldsymbol{x}\right)}{\partial y^{(l)}_i} u^{(l)}_{i}
\end{equation}
where $\boldsymbol{y}^{(l)}$ are the activations of the $l^{\text{th}}$ layer. Note that directional derivatives computed in the FG-A algorithm are defined as the sum over all neurons rather than all weights.
The activity-perturbed forward gradient $g^\mathrm{FG-A}\left(\boldsymbol{W}^{(l)}\right)$ is then:
\begin{equation}
	g^\mathrm{FG-A}\left(\boldsymbol{W}^{(l)}\right) := \left(d^\mathrm{FG-A} \boldsymbol{u}^{(l)}\right) \frac{\partial \boldsymbol{y}^{(l)}}{\partial \boldsymbol{W}^{(l)}}
\end{equation}
where $\nicefrac{\partial \boldsymbol{y}^{(l)}}{\partial \boldsymbol{W}^{(l)}}$ are local gradients computed by neurons with locally-available information. See Algorithm \ref{alg:fga} in Appendix for the full algorithm applied to a fully-connected DNN.
\par
This method reduces the number of derivatives to estimate since the number of neurons is considerably lower than the number of weights (see Table \ref{table:fc_depth} in Appendix for some examples). Consequently, the method leads to lower variance than FG-W \citep{scaling_forward_gradient_with_local_losses}.
\par
Activity-Perturbed FG demonstrated improvements over Weight-Perturbed FG on several benchmark datasets \citep{scaling_forward_gradient_with_local_losses} but still suffers from high variance as the number of neurons in DNNs remains large. To avoid this issue, \citeauthor{scaling_forward_gradient_with_local_losses} proposed the Local Greedy Activity-Perturbed Forward Gradient (LG-FG-A) method, which uses local loss functions to partition the gradient computation and decrease the number of derivatives to estimate. By adopting this local greedy strategy, LG-FG-A greatly improved the performance of Local MLP Mixers, a specific architecture that uses shallow multi-layer perceptrons to perform vision without having to use convolution. However, no results were reported for conventional fully-connected or convolutional neural networks, where the number of neurons per layer is larger than in MLP Mixers.

\subsection{Direct Feedback Alignment}

\begin{figure}[!tbp]
	\centering
	\begin{subfigure}[T]{0.5\textwidth}
		\includegraphics[width=\linewidth]{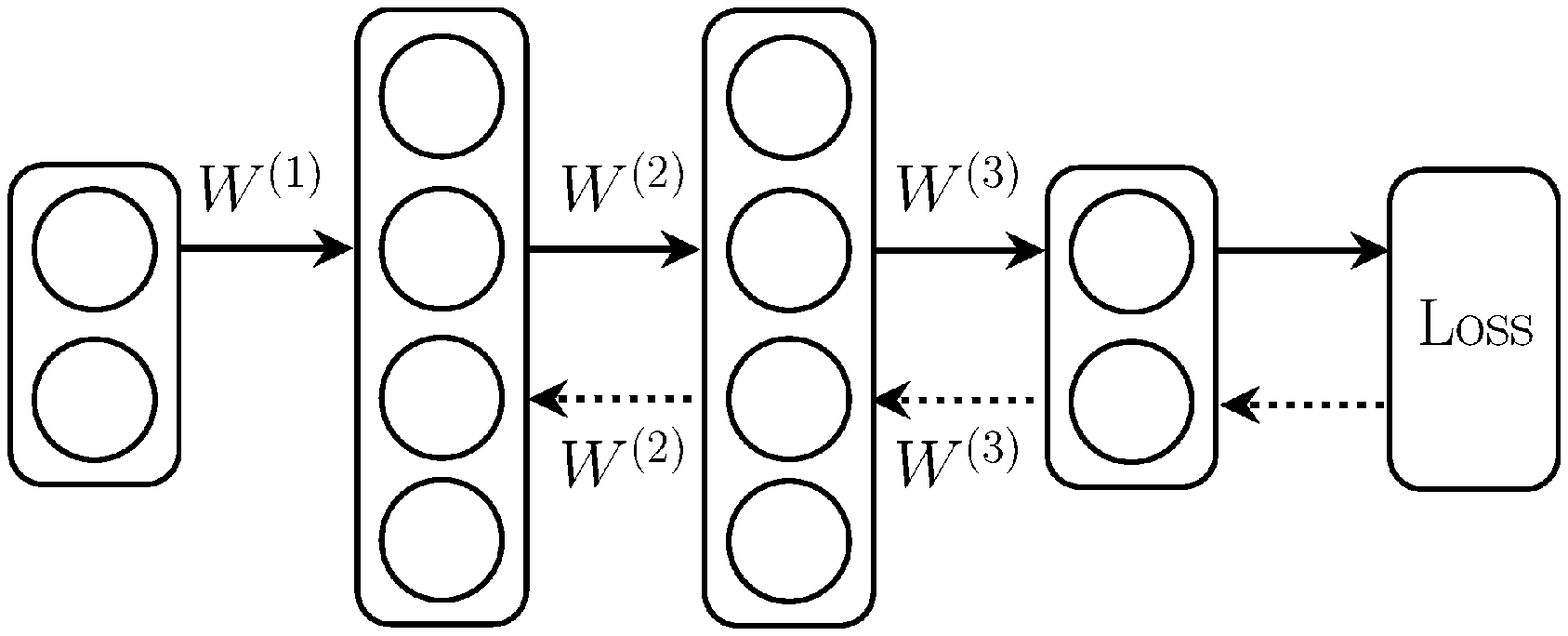}
		\caption{Error Backpropagation}
		\label{fig:error_backpropagation}
	\end{subfigure}%
	\begin{subfigure}[T]{0.5\textwidth}
		\includegraphics[width=\linewidth]{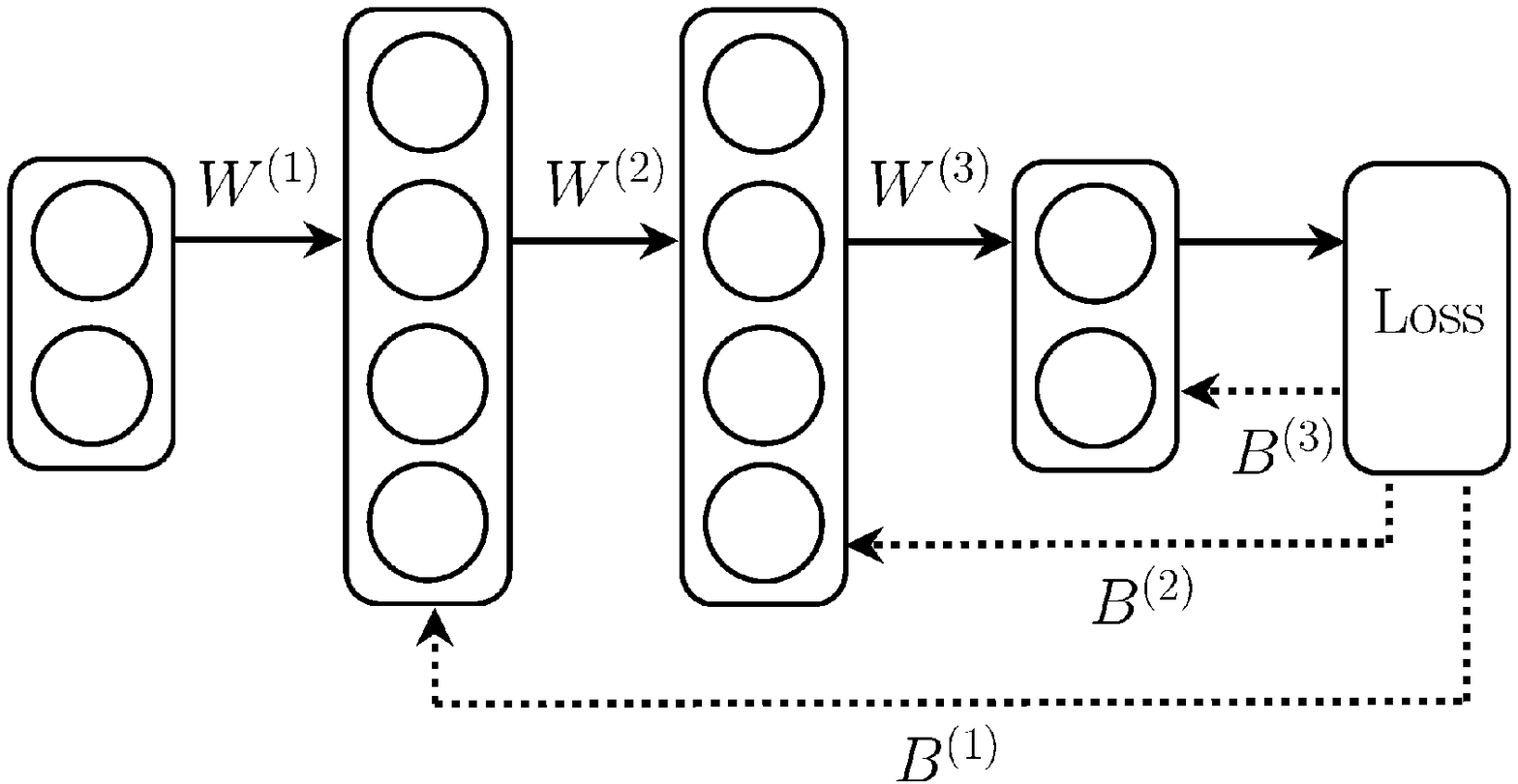}
		\caption{Direct Feedback Alignment}
		\label{fig:dfa}
	\end{subfigure}
	\caption{Illustrations of the error backpropagation (Figure \ref{fig:error_backpropagation}) and Direct Feedback Alignment (Figure \ref{fig:dfa}). Solid arrows represent forward paths and dotted arrows represent backpropagation paths.}
	\label{fig:backprop_vs_dfa}
\end{figure}

While BP relies on symmetric weights to propagate errors to hidden layers, it has been shown that weight symmetry is not mandatory to achieve learning \citep{random_fa}. For example, FA has proven that random fixed weights can also be used for backpropagating errors and still achieve learning \citep{random_fa}.
\par
DFA takes the idea of FA one step further by directly projecting output errors to hidden layers using fixed linear feedback connections \citep{random_dfa} (see Figure \ref{fig:backprop_vs_dfa}). Feedback matrices $B^{(l)} \in \mathbb{R}^{n_L \times n_l}$ replace the derivatives $\nicefrac{\partial \boldsymbol{y}^{(L)}}{\partial \boldsymbol{y}^{(l)}}$ of output neurons with respect to hidden neurons. The approximate gradient $g^\mathrm{DFA}\left(\boldsymbol{W}^{(l)}\right)$ for the weights of the hidden layer $l$ is then computed as follows:
\begin{equation}
	g^\mathrm{DFA}\left(\boldsymbol{W}^{(l)}\right) := \frac{\partial \mathcal{L}\left(\boldsymbol{x}\right)}{\partial \boldsymbol{y}^{(L)}} \boldsymbol{B}^{(l)} \frac{\partial \boldsymbol{y}^{(l)}}{\partial \boldsymbol{W}^{(l)}}
\end{equation}
In DFA, feedback matrices for hidden layers are chosen to be random and kept fixed during training. For the output layer, the identity matrix is used as no propagation of errors is required.
\par
The success of DFA depends on the alignment between the forward and feedback weights, which results in the alignment between the approximate and true gradient \citep{random_fa, random_dfa, align_then_memorise}. When the angle between these gradients is within 90 degrees, the direction of the update is descending \citep{random_fa, random_dfa}.
\par
DFA can scale to modern deep learning architectures such as Transformers \citep{dfa_scales_to_modern_dl} but is unable to train deep convolution layers \citep{principled_training_of_nn_with_dfa} and fails to learn challenging datasets such as CIFAR-100 or ImageNet without the use of transfer learning \citep{scalability_of_bio_motivated_dl, sparse_dfa}. However, recent methods to learn symmetric feedbacks such as the Direct Kolen-Pollack (DKP) algorithm \citep{learning_connections_in_dfa} showed promising results with convolutional neural networks due to improved gradient alignments.

\section{Method}

\begin{algorithm}
	\caption{Forward Direct Feedback Alignment algorithm with a fully-connected DNN.}
	\label{alg:fdfa}
	\begin{algorithmic}[1]
		\STATE {\bfseries Input:} Training data $\mathcal{D}$
		\STATE Randomly initialize $w^{(l)}_{ij}$ for all $l$, $i$ and $j$.
		\STATE Initialize $\boldsymbol{B}^{(l)}=\boldsymbol{0}$ for all $l$.
		\REPEAT
		\STATE \COMMENT{Inference (sequential)}
		\FORALL{$\boldsymbol{x}$ {\bfseries in} $\mathcal{D}$}
		\STATE $\boldsymbol{y}^{(0)} \leftarrow \boldsymbol{x}_s$
		\STATE $\boldsymbol{d}^{(0)} \leftarrow \boldsymbol{0}$
		\FOR{$l=1$ {\bfseries to} $L$}
		\STATE Sample $\boldsymbol{v}^{(l)} \sim \mathcal{N}\left(\boldsymbol{0}, \boldsymbol{I}\right)$
		\STATE $\boldsymbol{a}^{(l)} \leftarrow \boldsymbol{W}^{(l)} \boldsymbol{y}^{(l-1)}$
		\STATE $\boldsymbol{y}^{(l)} \leftarrow \sigma\left(\boldsymbol{a}^{(l)}\right)$
		\STATE $\boldsymbol{d}^{(l)} \leftarrow \left(\boldsymbol{W}^{(l)} \boldsymbol{d}^{(l-1)}\right) \odot \sigma^{\prime}\left(\boldsymbol{a}^{(l)}\right)$
		\IF{$l<L$}
		\STATE $\boldsymbol{d}^{(l)} \leftarrow \boldsymbol{d}^{(l)} + \boldsymbol{v}^{(l)}$
		\ENDIF
		\ENDFOR
		\STATE $\boldsymbol{e} \leftarrow \frac{\partial \mathcal{L}(\boldsymbol{x})}{\partial \boldsymbol{y}^{(L)}}$
		\STATE \COMMENT{Weights updates (parallel)}
		\STATE $\boldsymbol{W}^{(L)} \leftarrow \boldsymbol{W}^{(L)} - \lambda \; \boldsymbol{e} \otimes \boldsymbol{y}^{(L-1)}$
		\FOR{$l=1$ {\bfseries to} $L-1$}
		\STATE $\boldsymbol{B}^{(l)} \leftarrow \left(1 - \alpha\right) \boldsymbol{B}^{(l)} - \alpha \left(\boldsymbol{d}^{(L)} \otimes \boldsymbol{v}^{(l)}\right)$
		\STATE $\boldsymbol{W}^{(l)} \leftarrow \boldsymbol{W}^{(l)} - \lambda \left(\boldsymbol{e} \boldsymbol{B}^{(l)} \odot \sigma^\prime\left(\boldsymbol{a}^{(l)}\right)\right) \otimes \boldsymbol{y}^{(l-1)}$
		\ENDFOR
		\ENDFOR
		\UNTIL{$\mathbb{E}\left[\mathcal{L}(\boldsymbol{x})\right]<\epsilon$}
	\end{algorithmic}
\end{algorithm}

In this section, we describe our proposed Forward Direct Feedback Alignment (FDFA) algorithm which uses forward gradients to estimate derivatives between output and hidden neurons as direct feedback connections.
\par
Similarly to FG-A, we sample perturbation vectors $\boldsymbol{u}^{(l)} \in \mathbb{R}^{n_l}$ for each layer $l \leq L$ from a multivariate standard normal distribution. However, in contrast to FG-A, we use the directional derivatives that are computed at the output layer rather than at the loss function, such as:
\begin{equation}
	\boldsymbol{d}^\mathrm{FDFA} = \sum_{l=1}^{L-1} \sum_{i=1}^{n_l} \frac{\partial \boldsymbol{y}^{(L)}}{\partial y^{(l)}_i} u^{(l)}_{i}
\end{equation}
which produces a vector of directional derivatives. Note that only partial derivatives of the network outputs with respect to the activations of hidden neurons are considered, as represented by feedback connections in DFA. Output neurons are updated using a fixed identity feedback matrix as no error backpropagation is required.
\par
Rather than relying solely on the most recent forward gradient, as in FG-W and FG-A, it is possible to obtain a more accurate estimate of $\frac{\partial \boldsymbol{y}^{(L)}}{\partial \boldsymbol{y}^{(l)}}$ by averaging the forward gradient over the past training steps.
\par
Formally, we define an update rule for the direct feedback connections of DFA that performs an exponential average of the estimates, such as:
\begin{equation} \label{eq:fdfa_momentum}
	\boldsymbol{B}^{(l)} \leftarrow \left(1 - \alpha\right) \boldsymbol{B}^{(l)} + \alpha \; \boldsymbol{d}^{\mathrm{FDFA}} \otimes \boldsymbol{u}^{(l)}
\end{equation}
where $0 < \alpha < 1$ is the feedback learning rate. Equation \ref{eq:fdfa_momentum} can also be re-written in a form compatible with the stochastic gradient descent algorithm:
\begin{equation} \label{eq:fdfa_feedback_sgd_update}
	\boldsymbol{B}^{(l)} \leftarrow \boldsymbol{B}^{(l)} - \alpha \nabla \boldsymbol{B}^{(l)}
\end{equation}
where
\begin{equation}
	\nabla \boldsymbol{B}^{(l)} = \boldsymbol{B}^{(l)} - \boldsymbol{d}^{\mathrm{FDFA}} \otimes \boldsymbol{u}^{(l)}
\end{equation}
In essence, Equation \ref{eq:fdfa_feedback_sgd_update} minimizes the following Mean Squared Error (MSE) function:
\begin{equation}
	\mathcal{L}_B\left(\boldsymbol{x}\right) = \mathbb{E}\left[\sum_{l=1}^{L-1} \sum_{o=1}^{n_L} \sum	_{i=1}^{n_l} \left(B^{(l)}_{o,j} - d^{\mathrm{FDFA}}_{o} u^{(l)}_i\right)^2\right]
\end{equation}
where $d^{\mathrm{FDFA}}_{o} u^{(l)}_i$ is the target value for the optimized feedback connection $B^{(l)}_{o,j} $.
The global minimum of this loss function occurs at the point where all feedback connections are equal to the expected derivative between output and hidden neurons, such as:
\begin{equation}
	\begin{split}
		B^{(l)}_{o,i} = \mathbb{E}\left[d^{\mathrm{FDFA}}_{o} u^{(l)}_i\right] = \mathbb{E}\left[\frac{\partial y^{(L)}_o}{\partial y^{(l)}_i}\right]
	\end{split}
\end{equation}
for all output $o$ and all hidden neuron $i$ of all hidden layers $l<L$. Therefore, the FDFA algorithm is a dual optimization procedure where both loss functions $\mathcal{L}(\boldsymbol{x})$ and $\mathcal{L}_B(\boldsymbol{x})$ are minimized concurrently: weights are updated with direct feedbacks to minimize prediction errors and feedbacks connections converge towards the derivatives between output and hidden neurons. In the case where output neurons are linear, the penultimate feedback matrix becomes symmetric with the output weight matrix:
\begin{equation}
	\mathbb{E}\left[\boldsymbol{B}^{(L-1)}\right] = \mathbb{E}\left[\frac{\partial \boldsymbol{y}^{(L)}}{\partial \boldsymbol{y}^{(L-1)}}\right] = \boldsymbol{W}^{(L)}
\end{equation}
In this particular case, gradient estimates become mathematically equivalent to BP. For lower hidden layers, feedback matrices linearly approximate the non-linear derivatives between output and hidden neurons, which introduce a bias in the gradient estimates. This is because the feedback learning rule introduced in FDFA acts as momentum for these derivatives (see Equation \ref{eq:fdfa_momentum}), which is known to mitigate the effect of gradient variance on convergence at the cost of increased bias \citep{momentum_theoretical_insights}.
\par
Finally, the FDFA gradient estimate $g^\mathrm{FDFA}\left(\boldsymbol{W}^{(l)}\right)$ for the weights $\boldsymbol{W}^{(l)}$ of the hidden layer $l$ is computed as in DFA. Formally, output errors are linearly projected onto hidden neurons using the feedback matrix $B^{(l)}$, such as:
\begin{equation}
	g^\mathrm{FDFA}\left(\boldsymbol{W}^{(l)}\right) := \frac{\partial \mathcal{L}\left(\boldsymbol{x}\right)}{\partial \boldsymbol{y}^{(L)}} \boldsymbol{B}^{(l)} \frac{\partial \boldsymbol{y}^{(l)}}{\partial \boldsymbol{W}^{(l)}}
\end{equation}
The FDFA gradient estimate is then used to update the weights of the network with stochastic gradient descent, such as:
\begin{equation}
	\boldsymbol{W}^{(l)} \leftarrow \boldsymbol{W}^{(l)} - \lambda \; g^\mathrm{FDFA}\left(\boldsymbol{W}^{(l)}\right)
\end{equation}
where $\lambda > 0$ is a learning rate. Note that other gradient-based optimization techniques can also be applied for both feedback and forward weight updates in place of SGD, such as Adam \citep{adam}.
\par
The full algorithm applied to a fully connected DNN is given in Algorithm \ref{alg:fdfa}.

\section{Results}

We now present detailed theoretical and empirical results with our proposed FDFA algorithm and other local alternatives to BP. Details about our experimental settings are given in \ref{sec:experimental_settings}.

\subsection{Performance}

\begin{table*}[!tbp]
	\centering
	\caption{Performance of 4-layers fully connected DNNs trained on the MNIST, Fashion MNIST and CIFAR10 dataset.}
	\label{table:performances_fc}
	
	\footnotesize
	\begin{tabular}{|c|c|ccc|}
		\hline
		Method & Local & MNIST & Fashion MNIST & CIFAR10 \\
		\hline
		BP & \multirow{1}{*}{\xmark} & 98.46 $\pm$ 0.05\% & 89.99 $\pm$ 0.15\% & 56.92 $\pm$ 0.19\% \\ \hline
		FG-W & \multirow{6}{*}{\cmark} & 80.03 $\pm$ 0.62\% & 71.72 $\pm$ 0.63\% & 25.17 $\pm$ 0.32\% \\
		FG-A & & 83.40 $\pm$ 1.33\% & 77.62 $\pm$ 0.25\% & 35.53 $\pm$ 0.21\% \\
		LG-FG-A & & 86.89 $\pm$  0.90\% & 77.98 $\pm$ 0.21\% & 34.56 $\pm$ 0.34\% \\
		DFA & & 98.15 $\pm$ 0.05\% & 89.15 $\pm$ 0.09\% & 53.61 $\pm$ 0.16\% \\
		DKP & & 98.27 $\pm$ 0.05\% & 89.48 $\pm$ 0.09\% & 54.35 $\pm$ 0.21\% \\
		FDFA & & \textbf{98.32 $\pm$ 0.05\%} & \textbf{89.56 $\pm$ 0.11\%} & \textbf{54.97 $\pm$ 0.19\%} \\ \hline
	\end{tabular}
\end{table*}%
\begin{table*}[!tbp]
	\centering
	\caption{Performance of a shallow CNN (15C5-P2-40C5-P2-128-10 where 15C5 represents 15 5x5 convolutional layers and P2 represents a 2x2 max pooling layer) on the MNIST, Fashion MNIST and CIFAR10 datasets.}
	\label{table:performances_cnn}
	
	\footnotesize
	\begin{tabular}{|c|c|ccc|}
		\hline
		Method & Local & MNIST & Fashion MNIST & CIFAR10 \\
		\hline
		BP & \multirow{1}{*}{\xmark} & 99.32 $\pm$ 0.03\% & 92.10 $\pm$ 0.16\% & 68.11 $\pm$ 0.57\% \\ \hline
		FG-W & \multirow{5}{*}{\cmark} & 87.63 $\pm$ 1.20\% & 74.93 $\pm$ 0.55\% & 33.10 $\pm$ 0.37\% \\
		FG-A & & 96.63 $\pm$ 0.31\% & 82.71 $\pm$ 0.89\% & 42.51 $\pm$ 0.32\% \\
		LG-FG-A & & 95.42 $\pm$ 0.72\% & 79.96 $\pm$ 0.94\% & 37.74 $\pm$ 2.41\% \\
		DFA & & 98.80 $\pm$ 0.08\% & 89.69 $\pm$ 0.22\% & 58.14 $\pm$ 0.94\% \\
		DKP & & 99.08 $\pm$ 0.03\% & 90.39 $\pm$ 0.25\% & 60.30 $\pm$ 0.34\% \\
		FDFA & & \textbf{99.16 $\pm$ 0.03\%} & \textbf{91.54 $\pm$ 0.11\%} & \textbf{66.96 $\pm$ 0.70\%} \\ \hline
	\end{tabular}
\end{table*}%
\begin{table*}[!tbp]
	\centering
	\caption{Performance of AlexNet \citep{alexnet} trained on the CIFAR100 and Tiny ImageNet 200 datasets.}
	\label{table:performances_alexnet}
	
	\footnotesize
	\begin{tabular}{|c|c|cc|}
		\hline
		Method & Local & CIFAR100 & Tiny ImageNet \\
		\hline
		BP & \multirow{1}{*}{\xmark} & 60.43 $\pm$ 0.35\% & 40.55 $\pm$ 0.31\% \\ \hline
		FG-W & \multirow{5}{*}{\cmark} & 3.43 $\pm$ 0.43\% & 0.70 $\pm$ 0.09\% \\
		FG-A & & 15.84 $\pm$ 0.32\% & 3.52 $\pm$ 0.32\% \\
		LG-FG-A & & 15.27 $\pm$ 0.29\% & 2.90 $\pm$ 0.14\% \\
		DFA & & 35.75 $\pm$ 0.58\% & 17.47 $\pm$ 0.34\% \\
		DKP & & 49.15 $\pm$ 0.34\% & 25.36 $\pm$ 0.82\% \\
		FDFA & & \textbf{57.27 $\pm$ 0.11\%} & \textbf{36.47 $\pm$ 0.40\%} \\ \hline
	\end{tabular}
\end{table*}

We compared the performance of the proposed FDFA method with various local alternatives to BP related to our algorithm, including Weight-Perturbed Forward Gradient (FG-W), activity-perturbed Forward Gradient (FG-A), Local-Greedy activity-perturbed Forward Gradient (LG-FG-A), Random Direct Feedback Alignment (DFA) and the Direct Kolen-Pollack (DKP) algorithm. We evaluated each approach using fully-connected DNNs and Convolutional Neural Networks, including the AlexNet architecture \citep{alexnet}. We used several datasets widely adopted as standard benchmarks for training algorithm comparisons and presenting increasing levels of difficulty, namely MNIST, Fashion MNIST, CIFAR10, CIFAR100, and Tiny ImageNet 200. For each method, average test performance over 10 training runs is reported in Tables \ref{table:performances_fc}, \ref{table:performances_cnn}, and \ref{table:performances_alexnet}. Additional results with fully-connected DNNs of different depths are also given in Table \ref{table:performances_fc_depth} in \ref{seq:performances_fc_depth}.
\par
Overall, the FG-W algorithm achieves poor generalization compared to BP. More importantly, its performance significantly decreases with the size of the network and the complexity of the task. For example, the method is unable to converge with AlexNet on the CIFAR100 or Tiny ImageNet 200 datasets with 3.43\% and 0.70\% test accuracy respectively. The FG-A method slightly improves the generalization of forward gradients but still fails to match the performance of BP. The greedy approach used in LG-FG-A further improves performance when the number of neurons per layer is relatively low, as in fully-connected networks for MNIST and Fashion MNIST. However, the method does not perform as well as FG-A on networks that contain large layers such as convolution. This suggests, that LG-FG-A is most suitable for specific architectures where each local loss function sends error signals to a small number of neurons.
\par
DFA achieves performance closer to BP than forward gradient methods. However, it fails to scale to AlexNet on CIFAR100 and Tiny ImageNet 200, as observed in previous work \citep{principled_training_of_nn_with_dfa, scalability_of_bio_motivated_dl, sparse_dfa}. The performance of DFA is increased by the DKP algorithm but a gap still exists with BP, especially on difficult tasks such as CIFAR100 ad Tiny ImageNet 200. In contrast, our proposed FDFA method performs closer to BP than DFA and DKP on all benchmarked networks and datasets. For example, our method doubles the test accuracy of DFA and improves by at least 10\% the performance of DKP on Tiny ImageNet 200.

\subsection{Convergence}

\begin{figure*}[t]
	\centering
	
	\begin{tabular}{cc}
		\hspace{-1cm}\begin{minipage}{.45\textwidth}
			\includegraphics{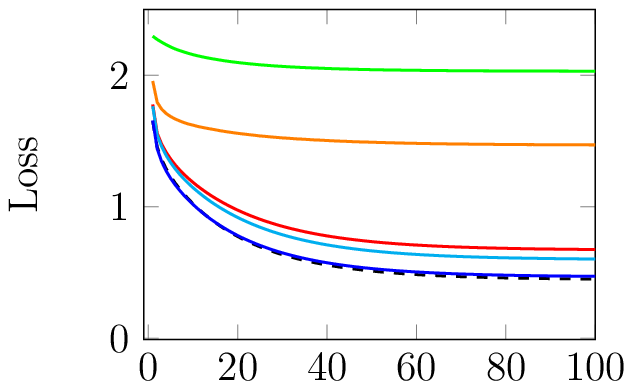}
		\end{minipage}
		&
		\hspace{0.5cm}\begin{minipage}{.45\textwidth}
			\includegraphics{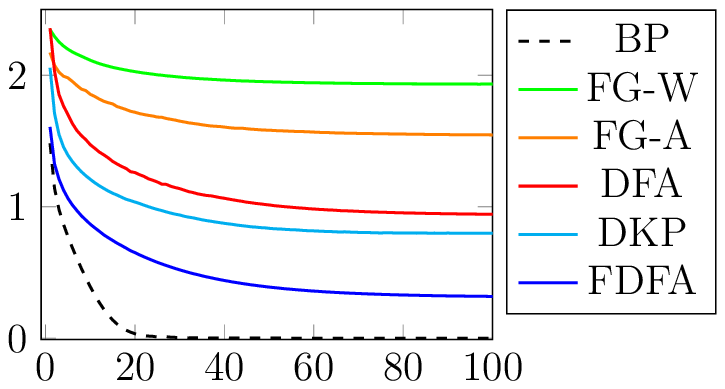}
		\end{minipage}
		\\
		\begin{minipage}{.4\textwidth}
			\leavevmode\subcaption{Fully-Connected}\label{fig:loss_cifar10_fc}
		\end{minipage}&
		\begin{minipage}{.4\textwidth}
			\leavevmode\subcaption{Convolution}\label{fig:loss_cifar10_conv}
		\end{minipage}
		\\
		\hspace{-1cm}\begin{minipage}{.45\textwidth}
			\includegraphics{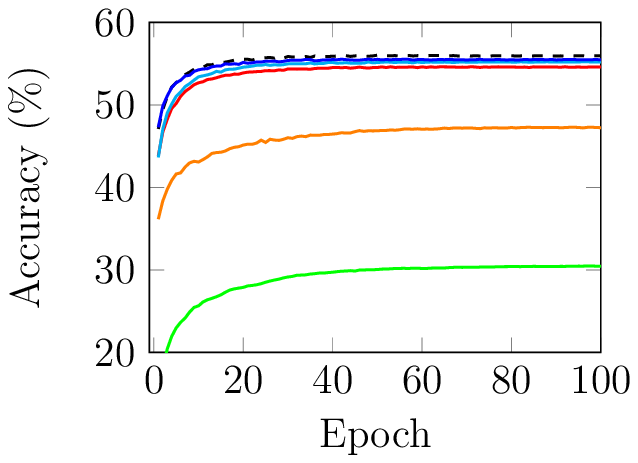}
		\end{minipage}
		&
		\hspace{0.2cm}\begin{minipage}{.45\textwidth}
			\includegraphics{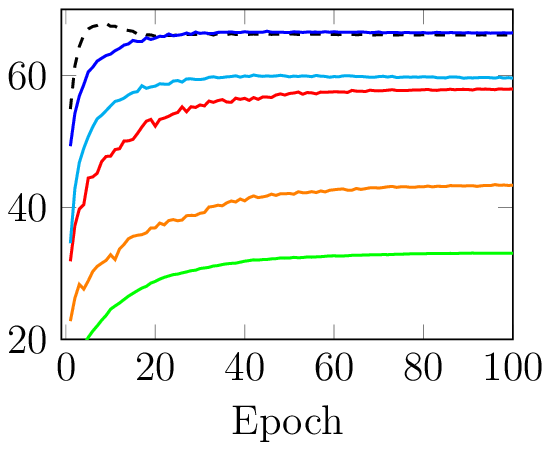}
		\end{minipage}
		\\
		\begin{minipage}{.4\textwidth}
			\leavevmode\subcaption{Fully-Connected}\label{fig:accuracy_cifar10_fc}
		\end{minipage}&
		\begin{minipage}{.4\textwidth}
			\leavevmode\subcaption{Convolution}\label{fig:accuracy_cifar10_conv}
		\end{minipage}
	\end{tabular}
	
	\caption{Training loss and test accuracy of a 2-layer fully-connected network (Figures \ref{fig:loss_cifar10_fc} and \ref{fig:accuracy_cifar10_fc}) and a CNN (Figures \ref{fig:loss_cifar10_conv} and \ref{fig:accuracy_cifar10_conv}) trained on the CIFAR10 dataset. FDFA has a similar convergence rate as BP on fully-connected networks. With CNNs, FDFA is not able to overfit the training data. However, our method has the highest convergence rate compared to FG-W, FG-A, and DFA.}
	\label{fig:loss_acc}
\end{figure*}

To evaluate the convergence improvements of our method, we measured the evolution of the training loss and test accuracy during the training process. As shown in Figure \ref{fig:loss_acc}, the FG-W algorithm converges slowly compared to both DFA and BP. Although FG-A slightly improves the convergence of FG-W, it was still unable to converge as quickly as BP and DFA. Both the DKP and FDFA algorithms showed better convergence rates than FG-W, FG-A, and DFA. However, DKP seems to be unable to reduce the loss as low as BP with convolutional layers. Finally, our proposed FDFA algorithm achieved a similar convergence rate as BP with fully connected networks and substantially improved the convergence rate of DFA and DKP with CNNs.

\subsection{Variance of Gradient Estimates}

To gain a deeper understanding of the reasons for the improvements observed in FDFA compared to FG-W and FG-A, we compare the theoretical variances of each method and empirically demonstrate the impact of gradient variance in the convergence of SDG.
\par
Following \citep{flipout} and \citep{scaling_forward_gradient_with_local_losses}, it can be shown that the variance of unbiased gradient estimates can be decomposed into two terms.
\begin{proposition} \citep{flipout, scaling_forward_gradient_with_local_losses}
	\label{col:variance_decomposition}
	Let $\boldsymbol{W}^{(1)} \in \mathbb{R}^{n_1, n_0}$ be the hidden weights of a two-layers fully-connected neural network. We denote by $\boldsymbol{x}$ the independent input samples and by $\boldsymbol{v}^{(1)}$ the independent random perturbations used to estimate the gradients of the hidden layer. The variance of a gradient estimate $g\left(w_{i,j}^{(1)}\right)$ for the hidden weight $w_{i,j}^{(1)}$ can be decomposed into two parts: 
	\begin{equation}
		\mathrm{Var}\left[g\left(w_{i,j}^{(1)}\right)\right] = z_1 + z_2
	\end{equation}
	where
	\begin{equation}
		\begin{split}
			z_1 := \underset{\boldsymbol{x}}{\mathrm{Var}}\left[\underset{\boldsymbol{v}^{(1)}}{\mathbb{E}}\left[g\left(w_{i,j}^{(1)}\right) \mid \boldsymbol{x}\right]\right]
		\end{split}
	\end{equation}
	and
	\begin{equation}
		\begin{split} 
			z_2 := \underset{\boldsymbol{x}}{\mathbb{E}}\left[\underset{\boldsymbol{v}^{(1)}}{\mathrm{Var}}\left[g\left(w_{i,j}^{(1)}\right) \mid \boldsymbol{x}\right]\right]
		\end{split}
	\end{equation}
\end{proposition}
The first term $z_1$ defined in Proposition \ref{col:variance_decomposition} corresponds to the gradient variance from data sampling. This term vanishes with the size of the batch in the case of mini-batch learning \citep{flipout, scaling_forward_gradient_with_local_losses}. The second term $z_2$ corresponds to the expected additional variance produced by the stochastic estimation of the gradient, which scales differently for each algorithm. In the case of BP and DFA, this term equals zero as both algorithms are deterministic. In \citep{scaling_forward_gradient_with_local_losses}, a third term $z_3$ was considered, which corresponds to the correlation between gradient estimates. However, this term is zero in the case where perturbations are independent \citep{scaling_forward_gradient_with_local_losses}.
\par
Next, we prove that the gradient estimation variance $z_2$ for the proposed FDFA algorithm vanishes as $\alpha$ tends towards zero.

\begin{proposition}
	\label{the:variance_fdfa}
	Let $\boldsymbol{W}^{(1)} \in \mathbb{R}^{n_1, n_0}$ be the hidden weights of a two-layers fully-connected neural network evaluated with an input sample $\boldsymbol{x} \in \mathbb{R}^{n_0}$. We denote by $g^\mathrm{FDFA}\left(w^{(1)}_{i,j}\right)$ the FDFA gradient estimate for the weight $w^{(l)}_{i,j}$ and assume that all the elements of the perturbation vector $\boldsymbol{u}^{(2)}$ for the activations of the output layer are 0. We also assume that all derivatives $\left(\nicefrac{\partial \mathcal{L}\left(\boldsymbol{x}\right)}{\partial w_{i,j}^{(l)}}\right)^2 \leq \beta$ are bounded and that the feedback matrix $\boldsymbol{B}^{(1)}$ converged to $\nicefrac{\partial \boldsymbol{y}^{(2)}}{\partial \boldsymbol{y}^{(1)}} = \boldsymbol{W}^{(1)}$, making the gradient estimates unbiased. If each element $u_{i}^{(1)} \sim \mathcal{N}(0, 1)$ of $\boldsymbol{u}^{(1)}$ follows a standard normal distribution, then: 
	\begin{equation}
		\underset{\boldsymbol{v}^{(1)}}{\mathrm{Var}}\left[g^\mathrm{FDFA}\left(w^{(1)}_{i,j}\right) \mid \boldsymbol{x}\right] = \alpha^2 \; \underset{\boldsymbol{v}^{(1)}}{\mathrm{Var}}\left[\sum_{k=1}^{n_1} \frac{\partial y^{(1)}_o}{\partial y_{k}^{(1)}} u_k^{(1)} u_i^{(1)}\right] \left(\frac{\partial y^{(1)}_i}{\partial w_{i,j}^{(1)}}\right)^2
	\end{equation}
	and
	\begin{equation}
		\lim_{\alpha \to 0}\underset{\boldsymbol{v}^{(1)}}{\mathrm{Var}}\left[g^\mathrm{FDFA}\left(w^{(1)}_{i,j}\right) \mid \boldsymbol{x}\right] = 0
	\end{equation}
\end{proposition}

\begin{proof}
	Starting from the gradient estimate $g^\mathrm{FG-A}\left(w^{(1)}_{i,j}\right)$, we have:
	\begin{equation}
		\begin{split}
			g^\mathrm{FDFA}\left(w^{(1)}_{i,j}\right) &= \sum_{o=1}^{n_2} \frac{\partial \mathcal{L}(\boldsymbol{x})}{\partial y_o^{(2)}} b_{o,i}^{(1)} \frac{\partial y^{(1)}_i}{\partial w_{i,j}^{(1)}} \\
			&= \sum_{o=1}^{n_2} \frac{\partial \mathcal{L}(\boldsymbol{x})}{\partial y_o^{(2)}} \left(\alpha \sum_{k=1}^{n_1} \frac{\partial y^{(1)}_o}{\partial y_{k}^{(1)}} u_k^{(1)} u_i^{(1)} + \left(1 - \alpha\right) \frac{\partial y_o^{(2)}}{\partial y_i^{(1)}}\right) \frac{\partial y^{(1)}_i}{\partial w_{i,j}^{(1)}} \\
			&= \left(1 - \alpha\right) \frac{\partial \mathcal{L}(\boldsymbol{x})}{\partial w_{i,j}^{(1)}} + \alpha \sum_{k=1}^{n_1} \frac{\partial y^{(1)}_o}{\partial y_{k}^{(1)}} u_k^{(1)} u_i^{(1)} \frac{\partial y^{(1)}_i}{\partial w_{i,j}^{(1)}}
		\end{split}
	\end{equation}
	Because both $\nicefrac{\partial \mathcal{L}(\boldsymbol{x})}{\partial w_{i,j}^{(1)}}$ and $\nicefrac{\partial y^{(1)}_i}{\partial w_{i,j}^{(1)}}$ are constants, the variance of $g^\mathrm{FDFA}\left(w^{(1)}_{i,j}\right)$ given an input sample $\boldsymbol{x}$ reduces to:
	\begin{equation}
		\begin{split}
			\underset{\boldsymbol{v}^{(1)}}{\mathrm{Var}}\left[g^\mathrm{FDFA}\left(w^{(1)}_{i,j}\right) \mid \boldsymbol{x}\right] =& \underset{\boldsymbol{v}^{(1)}}{\mathrm{Var}}\left[\left(1 - \alpha\right) \frac{\partial \mathcal{L}(\boldsymbol{x})}{\partial w_{i,j}^{(1)}} + \alpha \sum_{k=1}^{n_1} \frac{\partial y^{(1)}_o}{\partial y_{k}^{(1)}} u_k^{(1)} u_i^{(1)} \frac{\partial y^{(1)}_i}{\partial w_{i,j}^{(1)}} \mid \boldsymbol{x}\right] \\
			=& \underset{\boldsymbol{v}^{(1)}}{\mathrm{Var}}\left[\left(1 - \alpha\right) \frac{\partial \mathcal{L}(\boldsymbol{x})}{\partial w_{i,j}^{(1)}} \mid \boldsymbol{x}\right] \\
			& + \underset{\boldsymbol{v}^{(1)}}{\mathrm{Var}}\left[\alpha \sum_{k=1}^{n_1} \frac{\partial y^{(1)}_o}{\partial y_{k}^{(1)}} u_k^{(1)} u_i^{(1)} \frac{\partial y^{(1)}_i}{\partial w_{i,j}^{(1)}} \mid \boldsymbol{x}\right] \\
			=& \alpha^2 \; \underset{\boldsymbol{v}^{(1)}}{\mathrm{Var}}\left[\sum_{k=1}^{n_1} \frac{\partial y^{(1)}_o}{\partial y_{k}^{(1)}} u_k^{(1)} u_i^{(1)} \mid \boldsymbol{x}\right] \left(\frac{\partial y^{(1)}_i}{\partial w_{i,j}^{(1)}}\right)^2
		\end{split}
	\end{equation}
	Therefore:
	\begin{equation}
		\begin{split}
			\lim_{\alpha \to 0} \underset{\boldsymbol{v}^{(1)}}{\mathrm{Var}}\left[g^\mathrm{FDFA}\left(w^{(1)}_{i,j}\right) \mid \boldsymbol{x}\right] =& \lim_{\alpha \to 0} \alpha^2 \; \underset{\boldsymbol{v}^{(1)}}{\mathrm{Var}}\left[\sum_{k=1}^{n_1} \frac{\partial y^{(1)}_o}{\partial y_{k}^{(1)}} u_k^{(1)} u_i^{(1)} \mid \boldsymbol{x}\right] \left(\frac{\partial y^{(1)}_i}{\partial w_{i,j}^{(1)}}\right)^2 \\
			=& 0
		\end{split}
	\end{equation}
	which concludes the proof.
	
\end{proof}

\begin{figure}[t]
	\centering
	\begin{subfigure}[t]{0.45\textwidth}
		\begin{center}
			\hspace{-1cm}\includegraphics{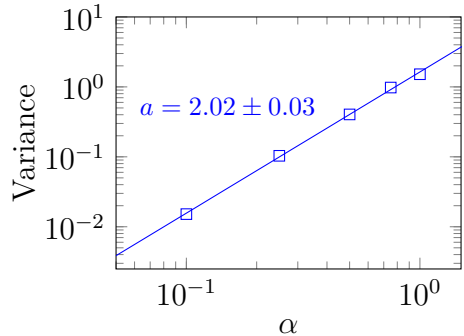}
		\end{center}
	\end{subfigure}
	\caption{The variance of FDFA gradient estimates as a function of the feedback learning rate $\alpha$ in a two-layer fully connected network. Each point was produced by computing the variance of gradient estimates over 10 iterations of the MNIST dataset. The blue line was fitted using linear regression. The slope $a$ and the asymptotic standard error is given with the same color. This figure shows that the variance of FDFA scales quadratically with $\alpha$. Note that both axes have a logarithmic scale. Therefore, the slope of the straight line on a logarithmc scale gives the power of the relationship on a linear scale.}
	\label{fig:variance_scaling_alpha}
\end{figure}

\begin{figure*}[t]
	\centering
	\begin{subfigure}[t]{0.45\textwidth}
		\begin{center}
			\includegraphics{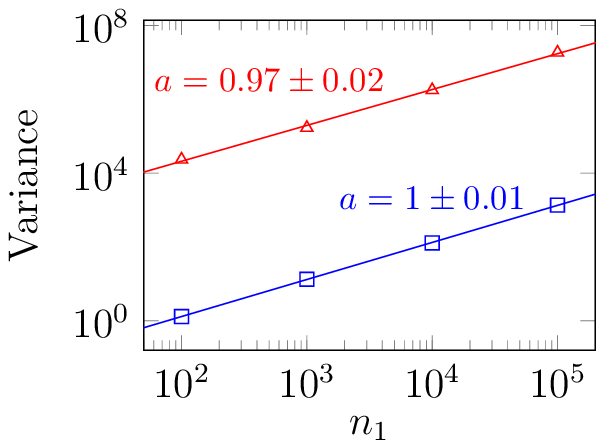}
			\caption{}
			\label{fig:variance_scaling_neuron}
		\end{center}
	\end{subfigure}%
	\begin{subfigure}[t]{0.45\textwidth}
		\begin{center}
			\vspace{0.25cm}
			\includegraphics{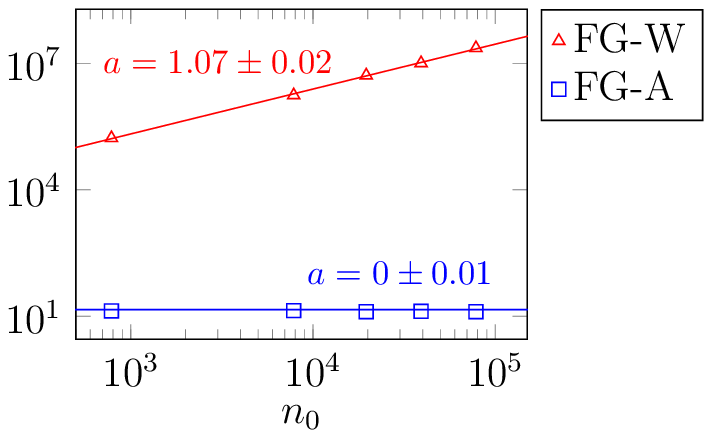}
			\caption{}
			\label{fig:variance_scaling_input}
		\end{center}
	\end{subfigure}
	\caption{Variance of the FG-W (red triangles) and FG-A (blue squares) gradient estimates as a function of the number of neurons $n_1$ (Figure \ref{fig:variance_scaling_neuron}) and number of inputs $n_0$ (Figure \ref{fig:variance_scaling_input}) in a two-layer fully connected network. Each point was produced by computing the variance of gradient estimates over 10 iterations of the MNIST dataset. The pixels of input images were duplicated, to increase the number of inputs $n_0$. The red and blue lines were fitted using linear regression. The slope $a$ and the asymptotic standard error of each line are given with the same color. These figures show that the variance of FG-W scales linearly with the number of neurons and inputs while the variance of FG-A only scales linearly with the number of neurons. Note that every axis has a logarithmic scale. Therefore, the slopes of straight lines relate to their exponents on linear scales.}
	\label{fig:variance_scaling}
\end{figure*}

We also derived the gradient estimation variance for the FG-W (Proposition \ref{the:variance_forward_gradient}) and FG-A (Proposition \ref{the:variance_fga}) algorithms in \ref{seq:variance_fg} and \ref{seq:variance_fga} respectively. Numerical verifications are provided in Figures \ref{fig:variance_scaling_alpha} and \ref{fig:variance_scaling}. Table \ref{table:variance_comparison} shows the comparison of these theoretical results.
\par
Proposition \ref{the:variance_fdfa} analytically shows that the gradient estimation variance of FDFA estimates quadratically vanishes as $\alpha$ tends towards zero (see Figure \ref{fig:variance_scaling_alpha} for numerical verifications). Therefore, by choosing values of $\alpha$ close to zero, the variance of FDFA estimates approaches the variance of the true gradient.
\par
In contrast, Propositions \ref{the:variance_forward_gradient} and \ref{the:variance_fga} show that the variance of FG-W and FG-A have different scaling with respect to the number of neurons and parameters in the network. FG-W scales linearly with the number of parameters and FG-A scales linearly with the number of neurons. This indicates that, in large DNNs, gradient estimates provided by the FG-W algorithm have a larger variance than FG-A estimates. However, as DNNs also contain large numbers of neurons, the variance of FG-A remains high.

\begin{table}[t]
	\centering
	
	\caption{Theoretical gradient estimation variance produced by the FG-W, FG-A and FDFA algorithm given a single input sample.}
	\label{table:variance_comparison}
	\footnotesize
	\begin{tabular}{|c|c|c|}
		\hline
		Algorithm & Local & Estimation Variance \\ \hline
		BP & \xmark & $0$ \\ \hline
		FG-W & \multirow{4}{*}{\cmark} & $O\left(n_1 n_0\right)$ \\
		FG-A & & $O\left(n_1\right)$ \\
		DFA & & $0$ \\
		FDFA ($\alpha \to 0$) & & $0$ \\ \hline
	\end{tabular}
\end{table}

\begin{figure}[t]
	\centering
	\begin{center}
		\begin{subfigure}[t]{0.6\textwidth}
			\begin{center}
				\includegraphics[width=\linewidth]{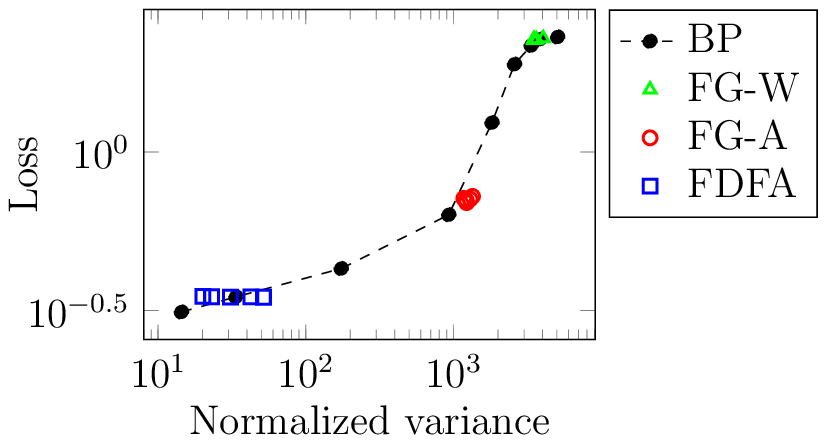}
			\end{center}
		\end{subfigure}
	\end{center}
	\caption{Correlation between the normalized variance of gradient estimates and the loss of a two-layer network with 1000 hidden neurons, following a single training epoch on the MNIST dataset. The variance of BP was artificially increased by adding Gaussian noise to the gradients to simulate the stochasticity of forward gradients. All gradient variances were normalized with the expected squared norm of the gradient estimates to ensure invariance with regard to the norm. Pairs of variance-loss for the FG-W, FG-A, and FDFA algorithms are represented in green, red, and blue, respectively. This figure shows that the differences in convergence are solely due to the variance of the gradient estimates.}
	\label{fig:variance_loss_relationship}
\end{figure}

To understand the impact of gradient variance on the convergence of SGD, we conducted an experiment where we measured the training loss of DNNs after one epoch as a function of variance. We simulated the variance of forward gradients by injecting Gaussian noise into the gradient computed with BP. By increasing the standard deviation of the injected noise, we artificially generated gradients with varying levels of variance. We then trained a two-layer DNN with these noisy gradients and measured the loss after one epoch of training on the MNIST dataset.
\par
Figure \ref{fig:variance_loss_relationship} illustrates that the variance strongly influences the training loss achieved after one epoch, as low-variance gradients tend to converge towards lower loss values compared to high-variance gradients. Hence, in this context, variance is the determining factor for convergence. Additionally, Figure \ref{fig:variance_loss_relationship} shows the variance and loss values associated with the FG-W, FG-A, and FDFA algorithms. Notably, all data points align with the line formed by the noisy BP. This shows that the differences in gradient variance are solely responsible for the differences in convergence among the FG-W, FG-A, and FDFA algorithms.

\subsection{Gradient Alignment in Feedback Methods}

\begin{figure*}[t]
	\centering
	\begin{subfigure}[t]{0.35\textwidth}
		\centering
		\hspace{-0.5cm}\includegraphics{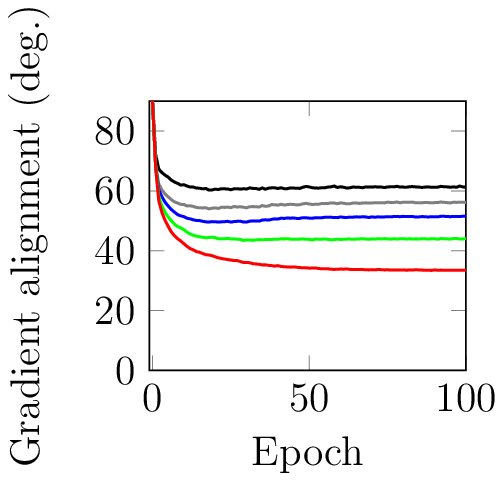}
		\caption{DFA}
		\label{fig:alignment_dfa}
	\end{subfigure}%
	\begin{subfigure}[t]{0.35\textwidth}
		\centering
		\includegraphics{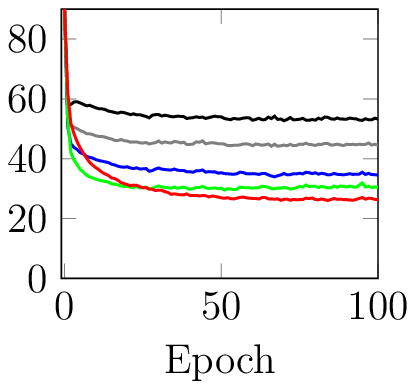}
		\caption{DKP}
		\label{fig:alignment_dkp}
	\end{subfigure}%
	\begin{subfigure}[t]{0.35\textwidth}
		\centering
		\includegraphics{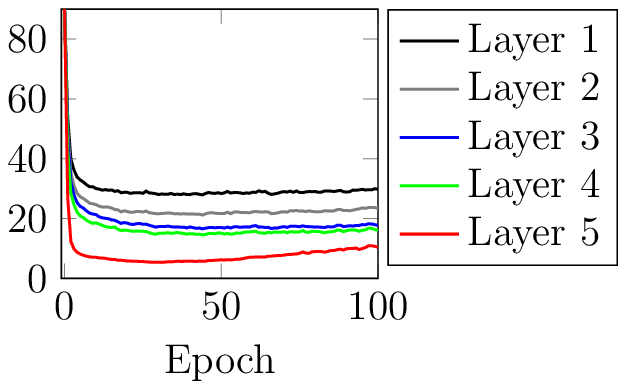}
		\caption{FDFA}
		\label{fig:alignment_fdfa}
	\end{subfigure}
	\caption{Layerwise alignment between gradient estimates and the true gradient computed using BP. These figures show that the proposed FDFA method (Figure \ref{fig:alignment_fdfa}) produces gradient estimates that better align with the true gradients than DFA (Figure \ref{fig:alignment_dfa}) which suggests improved descending directions.}
	\label{fig:alignment}
\end{figure*}

In our experiments, we observed that the proposed FDFA algorithm exhibits better convergence compared to DFA, despite having similar variance. This suggests that their gradient estimation variance alone does not fully explain the difference in convergence between these two algorithms.
\par
Both FDFA and DFA exhibit different degrees of biasedness. On the one hand, the FDFA algorithm introduces bias through its feedback learning rule, which acts as momentum for the derivatives between output and hidden neurons. On the other hand, DFA uses random feedback connections, resulting in strongly biased gradient estimates. Therefore, we conjecture that the disparity in convergence between DFA and the proposed FDFA algorithm can be attributed to their respective bias.
\par
To test this, we measured the bias of each method during training by recording the angle between the true gradient and the averaged gradient estimates produced by DFA, FDFA, and the DKP algorithm. This measure, often referred to as \textit{gradient alignment} \citep{align_then_memorise, learning_connections_in_dfa}, indicates the extent to which the expected gradient estimate deviates from the true gradient.
\par
Figure \ref{fig:alignment} shows the evolution of the layer-wise gradients alignment of each algorithm in a 5-layer fully connected DNN trained on the MNIST dataset during 100 epochs. This figure shows that the proposed FDFA method achieves faster alignment with the true gradient compared to DFA and DKP, which suggests earlier descending updates. Remarkably, the alignment of the gradient in FDFA is globally enhanced by 30 degrees in comparison to DFA and DKP. This significant improvement suggests that our method provides less biased gradient estimates. Thus, these findings strongly support the fact that the differences in convergence between DFA, DKP, and FDFA can be attributed to their respective levels of biasedness.

\section{Discussion}

The increasing size of DNNs, coupled with the emergence of low-powered neuromorphic hardware, has highlighted the need to explore alternative training methods that can overcome the limitations associated with BP. As a result, the exploration of local and parallel alternatives to BP has gained significant importance.
\par
In this work, we proposed the FDFA algorithm that combines the FG-A algorithm with DFA and momentum to train DNNs without relying on BP. The algorithm involves a dual optimization process where weights are updated with direct feedback connections to minimize classification errors, and derivatives are estimated as feedback using activity-perturbed forward gradients. Our experiments showed that the FDFA gradient estimate closely aligns with the true gradient, particularly in the last layers. However, the sole difference between the FDFA estimate and the true gradient are the feedback connections replacing the derivatives between outputs and hidden neurons. Hence, the convergence of feedback connections toward these derivatives is the only factor responsible for this gradients alignment. We can thus conclude that the proposed feedback learning rule is capable of learning the relevant derivatives to approximate BP. Morover, the averaging process introduced in our feedback learning rule acts as momentum for the derivatives estimates between output and hidden neurons. Our results demonstrated that the introduce momentum significantly reduces the gradient estimation variance compared to other forward gradient methods. Consequently, our method provides more accurate gradient estimates, leading to improved convergence and better performances on several benchmark datasets and architectures.
\par
Using our proposed method, feedback connections in the penultimate layer become equal to the output weights. This results in gradient estimates that are equivalent to BP. However, in deeper layers, feedback connections linearly approximate the non-linear derivatives between output and hidden neurons, introducing a bias. To address this bias, the FDFA algorithm could be adapted to Feedback Alignment, where feedback matrices replace the weights during the backward pass of BP, rather than direct connections between output and hidden layers. Although this modification would lower biasedness and better approximate BP without requiring weight transport, it would come at the cost of increased computation time due to backward locking. This highlights a particular tradeoff between computation cost and biasedness in FDFA. Future work could explore novel ways to leverage this biasedness without hindering the parallel aspect of FDFA.
\par
It is also important to note that, while both DFA and our proposed FDFA algorithm compute pseudo-gradients in a similar way, they differ fundamentally in their respective definitions of the feedback connections. In DFA, feedback connections are initialized randomly and remain fixed throughout training, thereby introducing a substantial bias into its pseudo-gradient. In contrast, our FDFA algorithms learns to represent the derivatives between output and hidden neurons as feedback connections. This distinction results in gradient estimates that align more closely with the steepest descent direction compared to DFA. This difference in biasedness was empirically verified in Figure \ref{fig:alignment} by measuring the angle between the gradient estimate of each algorithm and the true gradient computed with BP. As shown in Table \ref{table:variance_comparison}, DFA and FDFA have the same variance in the limit of small feedback learning rates (i.e. 0). Hence, their divergence lies solely in their respective biasness. We observed in our experiments that this difference in biasness consistently influences both the rate of convergence and the performance of DNNs on various benchmark datasets. Therefore, by providing an update direction closer to the direction of steepest descent, our proposed FDFA algorithm demonstrates the ability to converge faster and achieve greater performance than DFA.
\par
Finally, while methods relying on feedback connections showed better performance than forward gradients methods, the use of feedback matrices increases the number of weights to store in memory and ultimately poses a significant memory limitation, especially as networks grow larger. This highlights another important tradeoff between memory usage and performance. Future work could explore mechanisms such as sparse feedback matrices \citep{sparse_dfa} or reduced weight precision \citep{binary_dfa} to mitigate the memory impact of the proposed FDFA algorithm, further enhancing its practicality and scalability. Moreover, in the context of DNNs, local gradients of neurons are locally computed by differentiating the activations of neurons with respect to their weights. Similarly to DFA, the computation local change of weights is not restricted to the exact differentiation of neurons but can also be adapted to other local learning mechanisms. For example, the spikes fired by biologically plausible spiking neurons are known to be non-differentiable, making the computation of local gradients more challenging in Spiking Neural Networks. However, the FDFA could be adapted to SNNs by combining our feedback learning rule with locally-computed surrogate gradients or biologically plausible learning rules such as Spike Timing-Dependant Plasticity (STDP). Future work could thus explore the applicability of the FDFA algorithm to other local learning rules compatible with neuromorphic hardware.

\section{Conclusion}

The FDFA algorithm represents a promising local alternative to error backpropagation, effectively resolving backward locking and the weight transport problem. Its ability to approximate backpropagation with low variance not only opens new possibilities for the creation of efficient and scalable training algorithms but also holds significant importance in the domain of neuromorphic computing. By solely propagating information in a forward manner, the FDFA algorithm aligns with the online constraints of neuromorphic systems, presenting new prospects for developing algorithms specifically tailored to meet the requirements of these hardware. Therefore, the implications of our findings highlight the potential of FDFA as a promising direction of research for online learning on neuromorphic systems.

\bibliographystyle{elsarticle-harv}

\newpage

\appendix

\section{Experimental Settings} \label{sec:experimental_settings}

Each architecture uses the ReLU activation function in the hidden layers and linear activations in the output layer. For LG-FG-A, additional local linear outputs with their loss function were added after each layer $l < L-1$ to perform local greedy learning \citep{scaling_forward_gradient_with_local_losses}
In BP, FG-W, FG-A, LG-FG-A, DKP, and FDFA, forward weights were initialized with the uniform Kaiming initialization \citep{kaiming}. For FDFA, feedback connections were initialized to 0. For Random DFA, forward weights were initialized to 0 and feedback weights were drawn from a uniform Kaiming distribution and kept fixed during training. We removed the dropout layer in AlexNet as we found that it negatively affects feedback learning. Finally, we also added batch normalization \citep{batch_norm} after each convolutional layer of AlexNet to help training.
\par
We train each network over 100 epochs with Adam \citep{adam} and softmax cross-entropy loss functions. However, note that any differentiable loss functions can be used instead of the softmax cross entropy, such as the MSE. We also used Adam for the updates of feedback matrices in the FDFA and DKP algorithms. Adam was used because it is invariant to rescaling of the gradient \citep{adam}, making it a good optimization method to benchmark convergence with gradient estimators that exhibit different scales.
We used a learning rate of $\lambda=\alpha=10^{-4}$ and the default values of the parameters $\beta_1=0.9$, $\beta_2=0.999$ and $\epsilon=10^{-8}$ in Adam. Finally, no regularization or data augmentation has been applied. Finally, we used learning rate decay with a decay rate of $0.95$ after every epoch.
\par
In this work, all hyperparameters, including $\alpha$ and $\lambda$ were manually tuned. We found that, in many cases, better convergence behaviors were achieved if $\alpha \leq \lambda$. This motivated our choices for the learning rates of FDFA. However, hyperparameter optimization and learning rate scheduling could be used to further improve the convergence and performance of the method.

\newpage

\section{Performance of Fully-Connected Networks with Different Depths} \label{seq:performances_fc_depth}

\begin{table}[h]
	\centering
	\caption{Number of neurons and number of parameters in fully-connected DNNs with different depths.}
	\label{table:fc_depth}
	
	\scriptsize
	\renewcommand*{\arraystretch}{1.1}
	\begin{tabular}{|l|l|ccc|}
		\hline
		& Depth & MNIST & Fashion MNIST & CIFAR10 \\
		\hline
		\multirow{3}{*}{N. Neurons} & 2 Layers & 810 & 810 & 1010 \\
		& 3 Layers & 1610 & 1610 & 2010 \\
		& 4 Layers & 2410 & 2410 & 3010 \\ \hline
		\multirow{3}{*}{N. Params.} & 2 Layers & 636K & 636K & 3.08M \\
		& 3 Layers & 1.2M & 1.2M & 4.08M \\
		& 4 Layers & 1.9M & 1.9M & 5.08M \\ \hline
	\end{tabular}
\end{table}
\begin{table}[h]
	\centering
	\caption{Performance of fully-connected DNNs with different depths. LG-FG-A was not evaluated with two-layers DNNs as these networks are not deep enough to require greedy learning.}
	\label{table:performances_fc_depth}
	
	\scriptsize
	\renewcommand*{\arraystretch}{1.1}
	\begin{tabular}{|c|l|c|lll|}
		\hline
		Depth & Method & Local & MNIST & Fashion MNIST & CIFAR10 \\
		\hline
		\multirow{6}{*}{2 Layers} & BP & \multirow{1}{*}{\xmark} & 98.25 $\pm$ 0.04\% & 89.37 $\pm$ 0.09\% & 56.20 $\pm$ 0.12\% \\ \cline{2-6}
		& FG-W & \multirow{5}{*}{\cmark} & 85.87 $\pm$ 0.25\% & 77.72 $\pm$ 0.18\% & 30.47 $\pm$ 0.32\% \\
		& FG-A & & 93.39 $\pm$ 0.06\% & 84.73 $\pm$ 0.14\% & 47.39 $\pm$ 0.23\% \\
		& DFA & & 98.04 $\pm$ 0.07\% & 88.67 $\pm$ 0.12\% & 54.80 $\pm$ 0.22\% \\
		& DKP & & 98.12 $\pm$ 0.05\% & 89.09 $\pm$ 0.11\% & 55.41 $\pm$ 0.18\% \\
		& FDFA & & \textbf{98.21 $\pm$ 0.02\%} & \textbf{89.27 $\pm$ 0.07\%} & \textbf{55.74 $\pm$ 0.14\%} \\ \hline
		
		\multirow{7}{*}{3 Layers} & BP & \multirow{1}{*}{\xmark} & 98.39 $\pm$ 0.06\% & 90.01 $\pm$ 0.08\% & 56.66 $\pm$ 0.18\% \\ \cline{2-6}
		& FG-W & \multirow{6}{*}{\cmark} & 82.36 $\pm$ 0.48\% & 73.94 $\pm$ 0.57\% & 27.52 $\pm$ 0.34\% \\
		& FG-A & & 90.86 $\pm$ 0.17\% & 82.21 $\pm$ 0.13\% & 43.39 $\pm$ 0.14\% \\
		& LG-FG-A & & 91.64 $\pm$ 0.07\% & 82.41 $\pm$ 0.13\% & 43.08 $\pm$ 0.18\% \\
		& DFA & & 98.18 $\pm$ 0.06\% & 89.19 $\pm$ 0.13\% & 54.20 $\pm$ 0.11\% \\
		& DKP & & 98.28 $\pm$ 0.06\% & 89.63 $\pm$ 0.12\% & 55.15 $\pm$ 0.24\% \\
		& FDFA & & \textbf{98.30 $\pm$ 0.04\%} & \textbf{89.82 $\pm$ 0.13\%} & \textbf{55.70 $\pm$ 0.15\%} \\ \hline
		
		\multirow{7}{*}{4 Layers} & BP & \multirow{1}{*}{\xmark} & 98.46 $\pm$ 0.05\% & 89.99 $\pm$ 0.15\% & 56.92 $\pm$ 0.19\% \\ \cline{2-6}
		& FG-W & \multirow{6}{*}{\cmark} & 80.03 $\pm$ 0.62\% & 71.72 $\pm$ 0.63\% & 25.17 $\pm$ 0.32\% \\
		& FG-A & & 83.40 $\pm$ 1.33\% & 77.62 $\pm$ 0.25\% & 35.53 $\pm$ 0.21\% \\
		& LG-FG-A & & 86.89 $\pm$  0.90\% & 77.98 $\pm$ 0.21\% & 34.56 $\pm$ 0.34\% \\
		& DFA & & 98.15 $\pm$ 0.05\% & 89.15 $\pm$ 0.09\% & 53.61 $\pm$ 0.16\% \\
		& DKP & & 98.27 $\pm$ 0.05\% & 89.48 $\pm$ 0.09\% & 54.35 $\pm$ 0.21\% \\
		& FDFA & & \textbf{98.32 $\pm$ 0.05\%} & \textbf{89.56 $\pm$ 0.11\%} & \textbf{54.97 $\pm$ 0.19\%} \\ \hline
	\end{tabular}
\end{table}

\newpage

\section{Weight-Perturbed Forward Gradient Algorithm}

\begin{algorithm}[H]
	\caption{Weight-Perturbed Forward Gradient algorithm \citep{gradients_without_bp} with a fully-connected DNN.}
	\label{alg:forward_gradient}
	\begin{algorithmic}[1]
		\STATE {\bfseries Input:} Training data $\mathcal{D}$
		\STATE Randomly initialize $w^{(l)}_{ij}$ for all $l$, $i$ and $j$.
		\REPEAT
		\STATE \COMMENT{Inference (sequential)}
		\FOR{$\boldsymbol{x}$ {\bfseries in} $\mathcal{D}$}
		\STATE $\boldsymbol{y}^{(0)} \leftarrow \boldsymbol{x}_s$
		\STATE $\boldsymbol{d}^{(0)} \leftarrow \boldsymbol{0}$
		\FOR{$l=1$ {\bfseries to} $L$}
		\STATE Sample $\boldsymbol{V}^{(l)} \sim \mathcal{N}\left(\boldsymbol{0}, \boldsymbol{I}\right)$
		\STATE $\boldsymbol{a}^{(l)} \leftarrow \boldsymbol{W}^{(l)} \boldsymbol{y}^{(l-1)}$
		\STATE $\boldsymbol{y}^{(l)} \leftarrow f\left(\boldsymbol{a}^{(l)}\right)$
		\STATE $\boldsymbol{d}^{(l)} \leftarrow \left(\boldsymbol{W}^{(l)} \boldsymbol{d}^{(l-1)} + \boldsymbol{V}^{(l)} \boldsymbol{y}^{(l-1)}\right) \odot \sigma^{\prime}\left(\boldsymbol{a}^{(l)}\right)$
		\ENDFOR
		\STATE $d \leftarrow \frac{\partial \mathcal{L}(\boldsymbol{x})}{\partial \boldsymbol{y}^{(L)}} \boldsymbol{d}^{(L)}$
		\STATE \COMMENT{Weights updates (parallel)}
		\FOR{$l=1$ {\bfseries to} $L$}
		\STATE $\boldsymbol{W}^{(l)} \leftarrow \boldsymbol{W}^{(l)} - \lambda \boldsymbol{V}^{(l)} d$
		\ENDFOR
		\ENDFOR
		\UNTIL{$\mathbb{E}\left[\mathcal{L}(\boldsymbol{x})\right]<\epsilon$}
	\end{algorithmic}
\end{algorithm}

\section{Activity-Perturbed Forward Gradient Algorithm}

\begin{algorithm}[H]
	\caption{Activity-Perturbed Forward Gradient algorithm \citep{scaling_forward_gradient_with_local_losses} with a fully-connnected DNN.}
	\label{alg:fga}
	\begin{algorithmic}[1]
		\STATE {\bfseries Input:} Training data $\mathcal{D}$
		\STATE Randomly initialize $w^{(l)}_{ij}$ for all $l$, $i$ and $j$.
		\REPEAT
		\STATE \COMMENT{Inference (sequential)}
		\FORALL{$\boldsymbol{x}$ {\bfseries in} $\mathcal{D}$}
		\STATE $\boldsymbol{y}^{(0)} \leftarrow \boldsymbol{x}_s$
		\STATE $\boldsymbol{d}^{(0)} \leftarrow \boldsymbol{0}$
		\FOR{$l=1$ {\bfseries to} $L$}
		\STATE Sample $\boldsymbol{v}^{(l)} \sim \mathcal{N}\left(\boldsymbol{0}, \boldsymbol{I}\right)$
		\STATE $\boldsymbol{a}^{(l)} \leftarrow \boldsymbol{W}^{(l)} \boldsymbol{y}^{(l-1)}$
		\STATE $\boldsymbol{y}^{(l)} \leftarrow \sigma\left(\boldsymbol{a}^{(l)}\right)$
		\STATE $\boldsymbol{d}^{(l)} \leftarrow \left(\boldsymbol{W}^{(l)} \boldsymbol{d}^{(l-1)}\right) \odot \sigma^{\prime}\left(\boldsymbol{a}^{(l)}\right)$
		\IF{$l<L$}
		\STATE $\boldsymbol{d}^{(l)} \leftarrow \boldsymbol{d}^{(l)} + \boldsymbol{v}^{(l)}$
		\ENDIF
		\ENDFOR
		\STATE $d \leftarrow \frac{\partial \mathcal{L}(\boldsymbol{x})}{\partial \boldsymbol{y}^{(L)}} \boldsymbol{d}^{(L)}$
		\STATE \COMMENT{Weights updates (parallel)}
		\FOR{$l=1$ {\bfseries to} $L$}
		\STATE $\boldsymbol{W}^{(l)} \leftarrow \boldsymbol{W}^{(l)} - \lambda \left(d^{(L)} \boldsymbol{v}^{(l)} \odot \sigma^\prime\left(\boldsymbol{a}^{(l)}\right)\right) \otimes \boldsymbol{y}^{(l-1)}$
		\ENDFOR
		\ENDFOR
		\UNTIL{$\mathbb{E}\left[\mathcal{L}(\boldsymbol{x})\right]<\epsilon$}
	\end{algorithmic}
\end{algorithm}

\newpage

\section{Unbiasedness of Forward Gradients}

\begin{theorem}{Unbiasedness of Forward Gradients \citep{gradients_without_bp}}
	\label{the:forward_derivatives_unbiasedness}
	Let $\boldsymbol{x} \in \mathbb{R}^n$ be a vector of size $n$ and $\boldsymbol{v} \in \mathbb{R}^n$ be a random vector of $n$ independant variables. If $\boldsymbol{v} \sim \mathcal{N}\left(\boldsymbol{0}, \boldsymbol{I}\right)$ follows a multivariate standard normal distribution, then:
	\begin{equation}
		\mathbb{E}\left[\left(\boldsymbol{x} \cdot \boldsymbol{v}\right) \boldsymbol{x}\right] = \boldsymbol{x}
	\end{equation}
\end{theorem}

\begin{proof}
	
	Focusing on the $i^{\text{th}}$ element, we have:
	\begin{equation} \label{eq:expected_forward_derivative_decomposition}
		\begin{split}
			\mathbb{E}\left[\left(\boldsymbol{x} \cdot \boldsymbol{v}\right) v_i\right] &= \mathbb{E}\left[\sum_{j=1}^{n} x_j v_j v_i\right] \\
			&= \mathbb{E}\left[x_i v_i^2\right] + \sum_{\substack{j=1 \\ j \neq i}}^{n} \mathbb{E}\left[x_j v_j v_i\right] \\
			&= x_i \mathbb{E}\left[v_i^2\right] + \sum_{\substack{j=1 \\ j \neq i}}^{n} x_j \mathbb{E}\left[v_j\right] \mathbb{E}\left[v_i\right]
		\end{split}
	\end{equation}
	However, we know that $v_i \sim \mathcal{N}\left(0, 1\right)$. 
	Therefore, $\mathbb{E}\left[v_i\right] = 0$ and $\mathbb{E}\left[v_i^2\right] = \mathbb{E}\left[v_i\right]^2 + \mathrm{Var}\left[v_i\right] = 1$. Using these properties, Equation \ref{eq:expected_forward_derivative_decomposition} reduces to:
	\begin{equation}
		\underset{\boldsymbol{x}, \boldsymbol{v}}{\mathbb{E}}\left[\left(\boldsymbol{x} \cdot \boldsymbol{v}\right) v_i\right] = x_i
	\end{equation}
	and
	\begin{equation}
		\underset{\boldsymbol{x}, \boldsymbol{v}}{\mathbb{E}}\left[\left(\boldsymbol{x} \cdot \boldsymbol{v}\right) \boldsymbol{v}\right] = \boldsymbol{x}
	\end{equation}
	which concludes the proof.
	
\end{proof}

\newpage

\section{Variance of Forward Gradients}

\begin{lemma}
	\label{the:variance_forward_derivative}
	Let $\boldsymbol{x} \in \mathbb{R}^n$ be vector of size $n$ and $\boldsymbol{v} \in \mathbb{R}^n$ be a random vector of $n$ independant variables. If $\boldsymbol{v} \sim \mathcal{N}\left(\boldsymbol{0}, \boldsymbol{I}\right)$ follows a multivariate standard normal distribution, then:
	\begin{equation}
		\begin{split}
			\mathrm{Var}\left[\left(\boldsymbol{x} \cdot \boldsymbol{v}\right) v_i\right] =& x_i^2 + \left\| \boldsymbol{x} \right\|_2^2
		\end{split}
	\end{equation}
\end{lemma}

\begin{proof}
	Because the elements of $\boldsymbol{x}$ are considered as constants and all the elements of $\boldsymbol{v}$ are independant from each other, the variance of $\left(\boldsymbol{x} \cdot \boldsymbol{v}\right) v_i$ decomposes as follows:
	\begin{equation} \label{eq:der_var_decomposition} 
		\begin{split}
			\mathrm{Var}\left[\left(\boldsymbol{x} \cdot \boldsymbol{v}\right) v_i\right] =& \mathrm{Var}\left[\sum_{j=1}^{n} x_j v_j v_i\right] \\
			=& \sum_{j=1}^{n} x_j^2 \; \mathrm{Var}\left[v_j v_i\right]
		\end{split}
	\end{equation}
	We can show that, if $j \neq i$:
	\begin{equation} \label{eq:der_var_neq}
		\begin{split}
			\mathrm{Var}\left[v_j v_i\right] =& \mathbb{E}\left[\left(v_j v_i\right)^2\right] - \mathbb{E} \left[v_j v_i\right]^2 \\
			=& \mathbb{E}\left[v_i^2\right] \mathbb{E}\left[v_j^2\right] - \mathbb{E} \left[v_i\right]^2  \mathbb{E} \left[v_j\right]^2 \\
			=& 1
		\end{split}
	\end{equation}
	and if $j = i$:
	\begin{equation} \label{eq:der_var_eq}
		\begin{split}
			\mathrm{Var}\left[v_i^2\right] =& \mathbb{E}\left[\left( v_i^2\right)^2\right] - \mathbb{E}\left[v_i^2\right]^2 \\
			=& \mathbb{E} \left[v_i^4\right] - \mathbb{E} \left[v_i^2\right]^2 \\
			=& 2
		\end{split}
	\end{equation}
	as $\mathbb{E}\left[v_i^2\right]^2 = \mathrm{Var}\left[v_i\right]^2 = 1$ and $\mathbb{E}\left[v_i^4\right] = 3 \mathrm{Var} \left[v_i\right] = 3$. \\
	Therefore, by plugging Equations \ref{eq:der_var_neq} and \ref{eq:der_var_eq} into Equation \ref{eq:der_var_decomposition}, we find:
	\begin{equation}
		\begin{split}
			\mathrm{Var}\left[\left(\boldsymbol{x} \cdot \boldsymbol{v}\right) v_i\right] =& \sum_{j=1}^{n} x_j^2 \; \mathrm{Var}\left[v_j v_i\right] \\
			=& x_i^2 \mathrm{Var}\left[v_i^2\right] + \sum_{\substack{j=1\\j \neq i}}^{n} x_j^2 \mathrm{Var}\left[v_j v_i\right] \\
			=& 2 x_i^2 + \sum_{\substack{j=1\\j \neq i}}^{n} x_j^2 \\
			=& x_i^2 + \sum_{j=1}^{n} x_j^2 \\
			=& x_i^2 + \left\| \boldsymbol{x} \right\|_2^2
		\end{split}
	\end{equation}
	which concludes the proof.
\end{proof}

\newpage

\section{Variance of Weight-Perturbed Forward Gradients} \label{seq:variance_fg}

\begin{proposition}{Variance of Weight-Perturbed Forward Gradients} \\
	\label{the:variance_forward_gradient}
	Let $\boldsymbol{W}^{(1)} \in \mathbb{R}^{n_1, n_0}$ be the hidden weights of a two-layers fully-connected neural network evaluated with an input sample $\boldsymbol{x} \in \mathbb{R}^{n_0}$. We denote by $g^\mathrm{FG-W}\left(w^{(1)}_{i,j}\right)$ the weight-perturbed forward gradient for the weight $w^{(l)}_{i,j}$. We also assume that all derivatives $\left(\nicefrac{\partial \mathcal{L}\left(\boldsymbol{x}\right)}{\partial w_{i,j}^{(l)}}\right)^2 \leq \beta$ are bounded and that all the elements of the perturbation matrix $\boldsymbol{V}^{(2)}$ for the weights of the output layer are 0. If each element $v_{i,j}^{(1)} \sim \mathcal{N}(0, 1)$ of $\boldsymbol{V}^{(1)}$ follows a standard normal distribution, then:
	\begin{equation} \label{eq:fg_variance_scale}
		\begin{split}
			\underset{\boldsymbol{v}^{(1)}}{\mathrm{Var}}\left[g^\mathrm{FG-W}\left(w^{(1)}_{i,j}\right) \mid \boldsymbol{x}\right] =& \left(\frac{\partial \mathcal{L}(\boldsymbol{x})}{\partial w_{i,j}^{(1)}}\right)^2 + \left\|\frac{\partial \mathcal{L}(\boldsymbol{x})}{\partial \boldsymbol{W}^{(1)}}\right\|^2_2 \\
			=& O\left(n_1 n_0\right)
		\end{split}
	\end{equation}
	in the limit of large $n_1$ and large $n_0$.
\end{proposition}

\begin{proof}
	By application of Lemma \ref{the:variance_forward_derivative}, we know that:
	\begin{equation} \label{eq:fg_var_decomposition}
		\begin{split}
			\underset{\boldsymbol{v}^{(1)}}{\mathrm{Var}}\left[g^\mathrm{FG-W}\left(w^{(1)}_{i,j}\right) \mid \boldsymbol{x}\right] =& \underset{\boldsymbol{v}^{(1)}}{\mathrm{Var}}\left[\sum_{k=1}^{n_1} \sum_{l=1}^{n_0} \frac{\partial \mathcal{L}(\boldsymbol{x})}{\partial w^{(1)}_{k,l}} v_{k,l} \; v_{i,j}\right] \\
			=& \left(\frac{\partial \mathcal{L}(\boldsymbol{x})}{\partial w_{i,j}^{(1)}}\right)^2 + \left\|\frac{\partial \mathcal{L}(\boldsymbol{x})}{\partial \boldsymbol{W}^{(1)}}\right\|^2_2
		\end{split}
	\end{equation}
	However, in the limit of large $n_0$ and large $n_1$:
	\begin{equation}
		\left\| \frac{\partial \mathcal{L}(\boldsymbol{x})}{\partial \boldsymbol{W}^{(1)}}\right\|^2_2 = \sum_{i=1}^{n_1}\sum_{j=1}^{n_0} \left(\frac{\partial \mathcal{L}(\boldsymbol{x})}{\partial w^{(1)}_{i,j}}\right)^2 = O\left(n_1 n_0\right)
	\end{equation}
	Therefore, we conclude that:
	\begin{equation}
		\begin{split}
			\underset{\boldsymbol{v}^{(1)}}{\mathrm{Var}}\left[g^\mathrm{FG-W}\left(w^{(1)}_{i,j}\right) \mid \boldsymbol{x}\right] =& \; O\left(n_1 n_0\right)
		\end{split}
	\end{equation}
	
\end{proof}

\newpage

\section{Variance of Activity-Perturbed Forward Gradients} \label{seq:variance_fga}

\begin{proposition}{Variance of Activity-Perturbed Forward Gradients} \\
	\label{the:variance_fga}
	Let $\boldsymbol{W}^{(1)} \in \mathbb{R}^{n_1, n_0}$ be the hidden weights of a two-layers fully-connected neural network evaluated with an input sample $\boldsymbol{x} \in \mathbb{R}^{n_0}$. We denote by $g^\mathrm{FG-A}\left(w^{(1)}_{i,j}\right)$ the activity-perturbed forward gradient for the weight $w^{(l)}_{i,j}$. We also assume that all derivatives $\left(\nicefrac{\partial \mathcal{L}\left(\boldsymbol{x}\right)}{\partial w_{i,j}^{(l)}}\right)^2 \leq \beta$ are bounded and that all the elements of the perturbation vector $\boldsymbol{u}^{(2)}$ for the activations of the output layer are 0. If each element $u_{i}^{(1)} \sim \mathcal{N}(0, 1)$ of $\boldsymbol{u}^{(1)}$ follows a standard normal distribution, then:
	\begin{equation} \label{eq:fga_variance_scale}
		\begin{split}
			\underset{\boldsymbol{v}^{(1)}}{\mathrm{Var}}\left[g^\mathrm{FG-A}\left(w^{(1)}_{i,j}\right) \mid \boldsymbol{x}\right] =& \left[\left(\frac{\partial \mathcal{L}(\boldsymbol{x})}{\partial y^{(1)}_i}\right)^2 + \left\| \frac{\partial \mathcal{L}(\boldsymbol{x})}{\partial \boldsymbol{y}^{(1)}} \right\|^2_2\right] \left(\frac{\partial y^{(1)}_i}{\partial w_{i,j}^{(1)}}\right)^2 \\
			=& \; O\left(n_1\right)
		\end{split}
	\end{equation}
	in the limit of large $n_1$ and large $n_0$.
\end{proposition}

\begin{proof}
	
	Starting from the variance of $g^\mathrm{FG-A}\left(\boldsymbol{W}^{(1)}\right)$, we have:
	\begin{equation}
		\begin{split}
			\underset{\boldsymbol{v}^{(1)}}{\mathrm{Var}}\left[g^\mathrm{FG-A}\left(w^{(1)}_{i,j}\right) \mid \boldsymbol{x}\right] =& \underset{\boldsymbol{v}^{(1)}}{\mathrm{Var}} \left[\left(\sum_{k=1}^{n_1}\frac{\partial \mathcal{L}(\boldsymbol{x})}{\partial y^{(1)}_k} u_k^{(1)}\right) u_i^{(1)} \; \frac{\partial y^{(1)}_i}{\partial w_{i,j}^{(1)}}\right] \\
			=& \underset{\boldsymbol{v}^{(1)}}{\mathrm{Var}} \left[\sum_{k=1}^{n_1} \frac{\partial \mathcal{L}(\boldsymbol{x})}{\partial y^{(1)}_k} u_k^{(1)} u_i^{(1)} \mid \boldsymbol{x}\right] \left(\frac{\partial y^{(1)}_i}{\partial w_{i,j}^{(1)}}\right)^2
		\end{split}
	\end{equation}
	as the partial derivative $\frac{\partial y^{(1)}_i}{\partial w_{i,j}^{(1)}}$ is considered as constant. \\
	By applying Lemma \ref{the:variance_forward_derivative}, we obtain:
	\begin{equation}
		\begin{split}
			\underset{\boldsymbol{v}^{(1)}}{\mathrm{Var}}\left[g^\mathrm{FG-A}\left(w^{(1)}_{i,j}\right) \mid \boldsymbol{x}\right] =& \left[\left(\frac{\partial \mathcal{L}(\boldsymbol{x})}{\partial y^{(1)}_i}\right)^2 + \left\| \frac{\partial \mathcal{L}(\boldsymbol{x})}{\partial \boldsymbol{y}^{(1)}} \right\|^2_2\right] \left(\frac{\partial y^{(1)}_i}{\partial w_{i,j}^{(1)}}\right)^2 \\
			=& \left(\frac{\partial \mathcal{L}(\boldsymbol{x})}{\partial w_{i,j}^{(1)}}\right)^2 + \left\| \frac{\partial \mathcal{L}(\boldsymbol{x})}{\partial \boldsymbol{y}^{(1)}} \right\|^2_2 \left(\frac{\partial y^{(1)}_i}{\partial w_{i,j}^{(1)}}\right)^2
		\end{split}
	\end{equation}
	Finally, we can show that
	\begin{equation}
		\left\| \frac{\partial \mathcal{L}(\boldsymbol{x})}{\partial \boldsymbol{y}^{(1)}} \right\|^2_2 = \sum_{i=1}^{n_1} \left(\frac{\partial \mathcal{L}(\boldsymbol{x})}{\partial y^{(1)}_i}\right)^2 = O\left(n_1\right)
	\end{equation}
	Therefore:
	\begin{equation}
		\underset{\boldsymbol{v}^{(1)}}{\mathrm{Var}}\left[g^\mathrm{FG-A}\left(w^{(1)}_{i,j}\right) \mid \boldsymbol{x}\right] = O\left(n_1\right)
	\end{equation}
	
\end{proof}

\end{document}